\documentclass{article}

\usepackage[main, final]{neurips_2026}

\usepackage[utf8]{inputenc} 
\usepackage[T1]{fontenc}    
\usepackage{hyperref}       
\usepackage{url}            
\usepackage{booktabs}       
\usepackage{amsfonts}       
\usepackage{nicefrac}       
\usepackage{microtype}      
\usepackage{xcolor}         

\usepackage{amsmath}
\usepackage{amssymb}
\usepackage{mathtools}
\usepackage{amsthm}
\usepackage{placeins}
\usepackage{enumitem}
\setlist{topsep=1pt, itemsep=1pt, parsep=1pt}

\usepackage{hyperref}       
\usepackage{amsmath}        
\usepackage{url}            
\usepackage{booktabs}       
\usepackage{amsfonts}       
\usepackage{nicefrac}       
\usepackage{microtype}      
\usepackage{xcolor}         
\usepackage{graphicx}
\usepackage{amsmath}
\usepackage{algorithm}
\usepackage{algorithmic}
\usepackage{amsmath}
\usepackage{caption}
\usepackage{tabularx}
\usepackage{wrapfig}

\usepackage{subcaption}
\usepackage{mdframed}
\usepackage{newfloat}
\usepackage{placeins}

\DeclareFloatingEnvironment[
  fileext=lop,
  listname={List of Prompts},
  name=Prompt,
  placement=htp,
  within=none, 
]{prompt}

\newmdenv[
  backgroundcolor=gray!2,                  
  linecolor=gray!30,                       
  linewidth=0.5pt,                         
  roundcorner=5pt,                         
  font=\sffamily,                          
  frametitlefont=\sffamily\bfseries,       
  frametitlerule=false,                    
  frametitlealignment=\center,             
  innertopmargin=1em,                      
  innerbottommargin=1em,                   
  skipabove=1em,                           
  skipbelow=1em,                           
]{mymessagebox}

\title{Persona Generators: Generating Diverse Synthetic Personas for Arbitrary Contexts}

\author{%
  Davide Paglieri \\
  Google DeepMind \\
  \And
  Logan Cross \\
  Google DeepMind \\
  \And
  William A. Cunningham \\
  Google DeepMind \\
  \AND
  Joel Z. Leibo \\
  Google DeepMind \\
  \And
  Alexander Sasha Vezhnevets \\
  Google DeepMind \\
}

\begin{document}

\maketitle

\begin{abstract}
    Simulating human behavior with Large Language Models (LLMs) offers a scalable laboratory for social science and stress-testing AI products. However, limitations remain in aligning LLM outputs with the full breadth of human diversity. Current techniques for creating synthetic humans typically optimize for density matching, which often enforces behavioral uniformity and overlooks the rare, consequential outliers. We argue that robust simulation requires shifting the focus toward support coverage: ensuring synthetic populations span the entire landscape of possible human traits, opinions, and preferences. This paper introduces Persona Generators, functions that produce diverse synthetic populations for arbitrary contexts. We apply an iterative improvement loop based on AlphaEvolve, using LLMs as mutation operators to evolve our Persona Generator code rather than the personas themselves. The optimization process produces lightweight functions that automatically expand user provided scenario prompts into populations of diverse synthetic personas that are maximizing coverage along relevant diversity axes. We demonstrate that evolved generators substantially outperform existing baselines across six diversity metrics on held-out contexts. Furthermore, by successfully mitigating standard LLM mode collapse, our generated populations actually capture real human trait distributions better than baselines specifically designed to match human statistics.
\end{abstract}


\section{Introduction}

\begin{figure*}[ht]
    \centering
    \includegraphics[width=\linewidth]{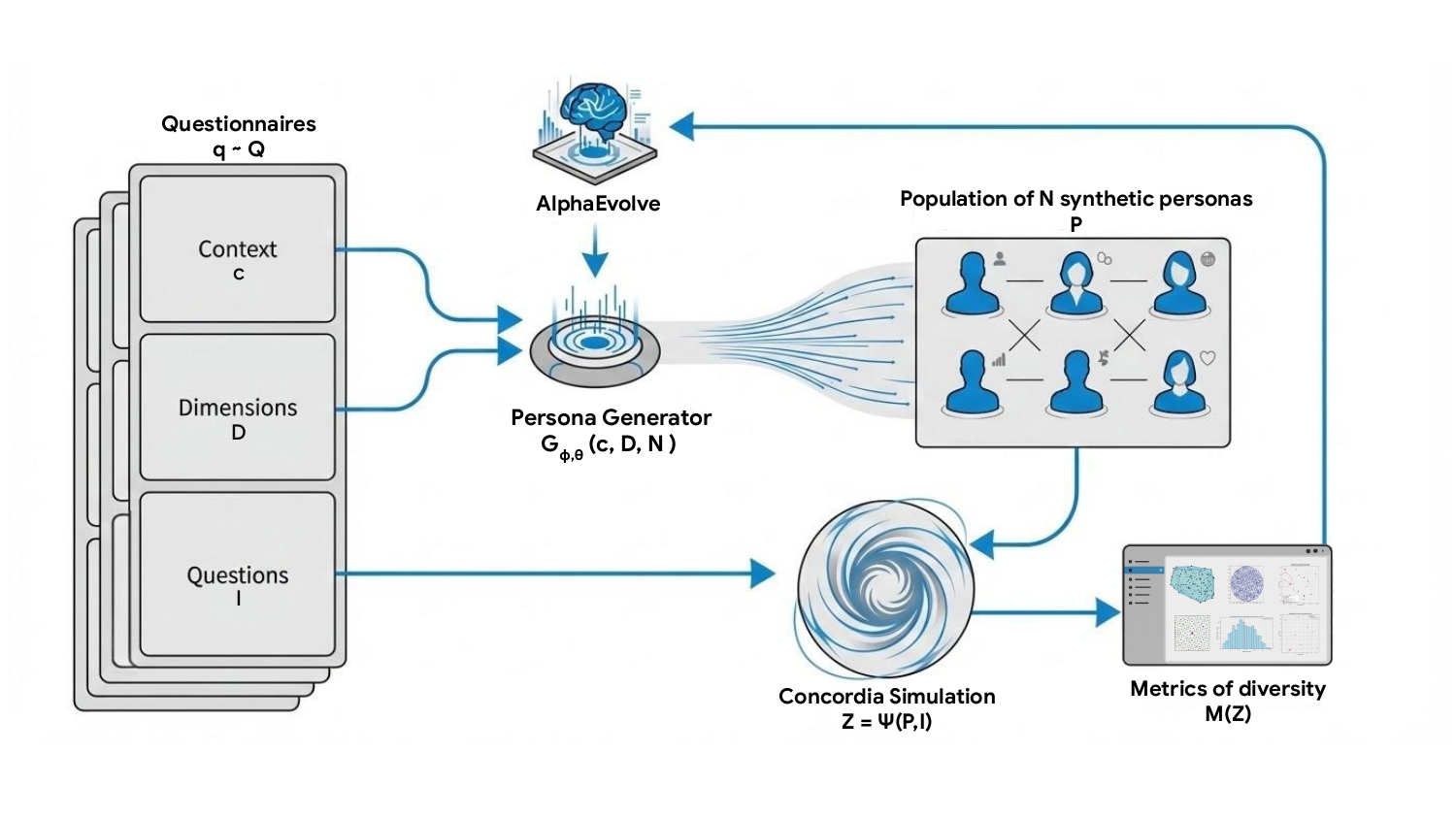}
    \caption{\textbf{Overview of the method.} First, we generate questionnaires containing specific contexts $c$, dimensions $\mathcal{D}$ (diversity axes) and questions $\mathcal{I}$ (items). The Persona Generator $G_{\phi, \theta}(c, \mathcal{D}, N)$ uses the context and diversity axes as inputs to create a population of synthetic personas $\mathcal{P}$. These personas are evaluated in Concordia simulations $Z = \Psi(\mathcal{P,I})$ where they answer the questionnaire items. We measure the diversity of their responses $\mathcal{M(Z)}$ and use AlphaEvolve to iteratively optimize the Persona Generator code $\phi$.}
    \vspace{-0.5cm}
    \label{diagram}
\end{figure*}

Simulating human populations is a promising research direction for answering social science questions in silico~\citep{anthis2025llm} and for evaluating human-AI interaction \citep{kim2023aligning,poole2025benchmarking}. Recent progress in Generative Agent-Based Modeling (GABM) has made it possible to simulate complex social interactions \citep{park2023generative}, societal response to COVID-19~\citep{kozlowski2024insilico}, behavioral patterns in surveys and economic games \citep{slumbers2025using,park2024generative,binz2025foundation}, social phenomena of status signaling~\cite{cross2026generative_cogsci}, and psychological phenomena of attitude change~\cite{matyas2026stabilising}. Whether modeling collective outcomes on social platforms or stress-testing personal health and coaching assistants, understanding the full breadth of potential human behaviors is critical for ensuring system robustness and safety. Yet, one major challenge in simulating human populations accurately is that standard LLMs often collapse onto a narrow subset of stereotypical or highly agreeable WEIRD (Western, Educated, Industrialized, Rich, Democratic) responses, rendering off-mode but consequential behaviors underrepresented \citep{bisbee2024synthetic, anthis2025llm, petrov2024limited}. Additionally, most existing work in GABM focuses on \emph{algorithmic fidelity}---how accurately synthetic personas reproduce observed human response patterns, often evaluated by matching the aggregate statistics or distributions derived from real data \citep{argyle2023out, park2024generative}. While effective for applications that prioritize the average user, density matching is fundamentally unsuited for stress-testing or forecasting in speculative scenarios—such as societal response to novel technology (e.g. AGI)—where exploring the full range of possible human responses is more critical than identifying the most probable one.

In this paper, we propose support coverage as a primary objective: creating populations that capture the \textit{full support} of possible attitudes and preferences. Leaving the long tail underexplored can create a false sense of robustness; in many stress-testing settings, it is the outliers—such as rare users with extreme behaviors—and not the average user, that drive critical failures. The density matching objective is not sensitive to these cases since their share of the population is low. 

Eliciting such diversity from LLM-based personas, however, is far from trivial. Naive prompting leads to mode collapse, stereotypical outputs, and systematic biases, partly a byproduct of Reinforcement Learning from Human Feedback (RLHF) tuning, even when explicit instructions for diversity are provided \citep{santurkar2023whose, li2025llm}. Our experiments show that simply asking an LLM to "generate diverse personas" typically yields populations clustered around stereotypical responses, failing to cover extreme or unusual trait combinations.

Rather than constructing a single fixed population of synthetic users, we propose to learn a reusable Persona Generator: a function capable of producing diverse synthetic personas on demand for any arbitrary context from a single scenario prompt (e.g. "clients of food delivery apps in London"). This generator must handle the immense variety of potential scenarios that we may care about. Importantly, we aim to optimize a generally capable Persona Generator, which samples and constructs a population of personas (see Fig.~\ref{persona_generator_image}).

Starting from a short scenario prompt, we first expand the context into a structured questionnaire that defines a set of diversity axes (e.g. dietary preferences, budget). A Persona Generator then produces a population of synthetic individuals intended to span all possible traits, opinions, and preferences defined by those axes. To overcome the difficulties of achieving diversity through standard prompting, we frame the task as an optimization problem over the Persona Generator's code, including prompt templates and sampling logic, using an evolutionary search loop powered by AlphaEvolve \citep{novikov2025alphaevolve}.

The resulting Persona Generator is lightweight and efficient, enabling rapid, one-shot population synthesis for downstream applications. This separation between a costly training phase and a cheap inference phase makes Persona Generators practical for repeated use across domains, even when the original optimization context differs from the deployment setting. Crucially, we find that maximizing support coverage does not come at the expense of realism. By explicitly fighting LLM mode collapse, our approach generates populations that actually match the true variance of real human trait distributions better than baselines specifically grounded in demographic data. Ultimately if the full support is covered, one can always subsequently re-sample the population to match any specific target density.

Our contributions are summarized as follows:

\begin{itemize}
    \item We formalize the problem of synthetic persona generation as a diversity maximization task over trait and preference embeddings, explicitly shifting the objective from algorithmic fidelity (matching specific individuals) to support coverage (spanning the space of possible traits, opinions, and preferences).
    \item We propose a novel Persona Generator function with a two-stage scalable architecture that separates population-level diversity decisions from per-persona background expansions, enabling both control and efficiency, coupled with a scalable pipeline that uses LLMs to generate questionnaires, simulate interactions, and evolve code.
    \item We demonstrate that LLM-driven evolution can discover novel Persona Generator functions that substantially outperform different baselines in coverage and diversity metrics. Furthermore, we show that by successfully mitigating LLM mode collapse, our generated personas actually capture real-world human trait distributions better than baselines specifically intended to mirror human statistics.
\end{itemize}


\section{Evolving Persona Generators}

We introduce a method for learning Persona Generators: functions capable of producing diverse populations of synthetic personas for \textit{any} arbitrary context. Our full pipeline is illustrated in Figure \ref{diagram}, and this section will walk through each of its core components.

The process starts by generating a set of evaluation tasks. As we will detail in Section \ref{questionnaire_generator}, we create synthetic questionnaires ($q$), each comprising a context ($c$), axes of diversity ($D$), and survey items ($I$). These questionnaires serve as the benchmark for evaluating the diversity of our generated populations.

The central component is the Persona Generator ($G_{\phi,\theta}$), which takes a questionnaire's context and diversity axes to produce a population of personas ($P$). The architecture of this generator, shown in Figure \ref{persona_generator_image}, follows a two-stage process to efficiently generate diverse individuals. We describe the initial implementations of the generators in Section \ref{initial_persona_generators}.

To evaluate a population, each persona is placed in a Concordia simulation where they answer the questionnaire items (see Section \ref{simulations}). This process maps the population of text-based personas to a set of numerical response embeddings ($Z$). We then quantify how well these embeddings cover the target space using several diversity metrics ($M(Z)$), which are defined in Section \ref{diversity_metrics}.

Finally, we use this evaluation as the fitness function in an evolutionary search loop driven by AlphaEvolve (Section \ref{alphaevolve_loop}) \citep{novikov2025alphaevolve}. This loop iteratively mutates the code ($\phi$) of the Persona Generator to discover new versions that maximize the diversity of the personas they produce.

\subsection{Problem Formulation}

We formalize the problem of generating populations of diverse synthetic personas as follows. Let $\mathcal{Q}$ be a distribution of questionnaires, with each questionnaire $q \sim \mathcal{Q}$ defined as a tuple $q = (c, \mathcal{D}, \mathcal{I})$, where $c$ is a textual context, $\mathcal{D} = \{d_1, \dots, d_K\}$ is a set of $K$ axes of diversity or dimensions, and $\mathcal{I}$ is a set of questions (items).

A Persona Generator is a function $G_{\phi, \theta}$ parameterized by code $\phi$ and a fixed large language model $\theta$ which executes API calls within $\phi$. The generator takes the context $c$, dimensions $\mathcal{D}$ and a target population size $N$ as inputs and produces a population of synthetic personas $\mathcal{P} = \{p_1, \dots, p_N\} = G_{\phi, \theta}(c, \mathcal{D}, N)$, where $p_{i}$ are text-based descriptions of synthetic individuals.

We evaluate the diversity of the population $\mathcal{P}$ with respect to the axes of diversity $\mathcal{D}$ of a questionnaire $q$ by asking each persona the items $\mathcal{I}$ in a simulation. We formalize the simulation as a mapping function $\psi$, that maps each persona to a response embedding by aggregating their answers into a score vector $\mathbf{z}_i \in \mathbb{R}^{|\mathcal{D}|}$, with $\mathbf{z}_i = \psi(p_i, \mathcal{I})$.

We define the population embedding $\mathcal{Z} = \Psi(\mathcal{P}, \mathcal{I})$ as the element-wise application of $\psi$ to $\mathcal{P}$. Finally, we define some diversity metrics $\mathcal{M}(\mathcal{Z})$, that quantify how well the population covers the search space spanned by $\mathcal{D}$. Our goal is to find the code $\phi^*$ that maximizes these metrics:
\begin{equation}
    \phi^* = \operatorname*{argmax}_{\phi} \mathbb{E}_{(c, \mathcal{D}, \mathcal{I}) \sim \mathcal{Q}} \left[ \mathcal{M}\left( \Psi(G_{\phi, \theta}(c, \mathcal{D}, N), \mathcal{I}) \right) \right]
\end{equation}
Because the true support of human attitudes is unknown and need not be convex, and exact coverage of a continuous support is impossible with a finite population $N$, Equation~(1) operationalizes support coverage over the bounded questionnaire-response domain defined by the fixed axes $\mathcal{D}$.

\subsection{Questionnaire Generator}
\label{questionnaire_generator}

We build an automated questionnaire generator using few-shot prompting to create questionnaires $q = (c, \mathcal{D}, \mathcal{I})$. As demonstrations, we use four well-established and publicly available questionnaires: the Big Five Inventory (BFI) \citep{john1999bigfive}, the Depression Anxiety Stress Scale (DASS) \citep{lovibond1995structure}, the Social Value Orientation (SVO) scale \citep{messick1968motivational, murphy2011measuring}, and the Need for Closure Scale (NFCS) \citep{webster1994individual}. For each of the above we extract a short context $c$ describing what the questionnaire is about, the axes of diversity $\mathcal{D}$, and the list of questions $\mathcal{I}$ grouped by axis. We then convert them into python code compatible with Concordia \citep{vezhnevets2023generative}, and use them as few-shot examples for Gemini 2.5 Pro \citep{comanici2025gemini}.

To generate a new questionnaire, we provide a very short description $\hat{c}$ of what the questionnaire should measure and prompt the model to produce the full questionnaire in two steps. First, the model expands the short description into a more detailed context $c$ and proposes relevant axes $\mathcal{D}$ (typically $K=2$ or $3$). It then generates items $\mathcal{I}$ grouped by axis, using a 5-point Likert scale format \citep{likert1932technique}. We generate 50 distinct questionnaires, split into 30 training, 10 validation, and 10 test sets. Appendix~\ref{generated_questionnaires} provides further details and examples.

\subsection{Persona Generators}

\begin{figure*}[t]
    \centering
    \includegraphics[width=\linewidth]{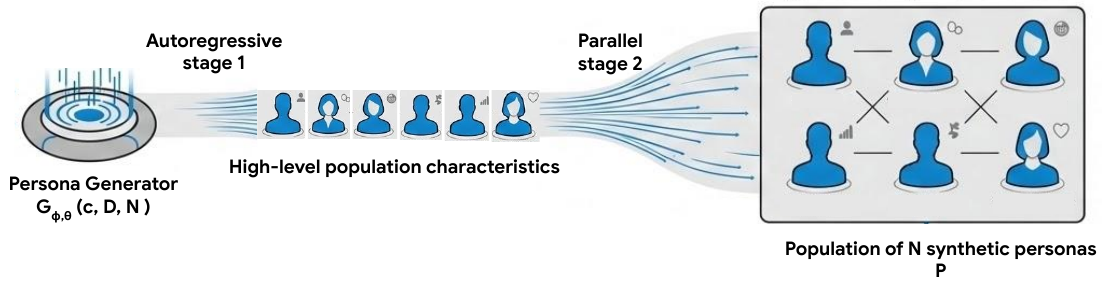}
    \caption{\textbf{Two stage Persona Generator.} The Persona Generator $G_{\phi, \theta}(c, \mathcal{D}, N)$ works in two stages: the autoregressive stage 1 generates high-level descriptors for each persona, then the parallel stage 2 expands the high level descriptions of each persona by generating additional details.}
    \label{persona_generator_image}
\end{figure*}

\label{initial_persona_generators}

The Persona Generator $G_{\phi, \theta}$, and specifically its code $\phi$, is the core component that we aim to optimize. Persona Generators take as input a context $c$, axes of diversity $\mathcal{D}$, and a target population size $N$, and produce a population of text-based personas $\mathcal{P}$. The code $\phi$ that handles the generation follows a two-stage pipeline. \begin{enumerate}
    \item \textbf{Autoregressive Stage 1:} The generator prompts the LLM $\theta$ to autoregressively produce high-level descriptors $\hat{p_i}$ for each persona. Critically it decides at a high-level where each individual is positioned along the specified diversity axes $\mathcal{D}$.
    \item \textbf{Parallel Stage 2:} Each high-level descriptor $\hat{p_i}$ is expanded into a complete persona $p_i$ by generating additional contextual details and specific traits.
\end{enumerate}

Stage~1 is responsible for shaping population-level diversity, while stage~2 is executed in parallel and mainly serves efficiency. A diagram of the Persona Generator can be seen in figure \ref{persona_generator_image}

We initialize the evolutionary search with three initial implementations of $\phi$. The first is a baseline formative memory generator, adapted from the default Concordia Persona Generator \citep{vezhnevets2023generative}. The second modifies stage~1 to generate personas in smaller autoregressive batches, reducing dependencies between consecutive individuals. The third introduces a quasi-random Monte Carlo sampling scheme to select continuous positions along each diversity axis. These sampled positions are then translated into high-level traits descriptions $\hat{p}_{i}$ by an LLM, before being expanded into full personas $p_i$ in stage~2. All text generation in the Persona Generators is performed by gemma-3-27b-it \citep{team2025gemma} (represented by $\theta$).
\vspace{-0.25cm}
\subsection{Concordia Simulations}
\label{simulations}

The simulation function $\Psi$ maps the population of textual personas $\mathcal{P}$ into a set of behavioral vectors $\mathcal{Z}$ by asking the questions $\mathcal{I}$ from the questionnaire. We implement this process using the Concordia library. For each persona $p_i \in \mathcal{P}$, we instantiate a basic Concordia agent that acts based on the logic of appropriateness \citep{march2011logic, leibo2024theory}. When presented with a question from $\mathcal{I}$, we use the LLM $\theta$ (here gemma-3-27b-it) to role-play the persona’s response by asking it to reply to the three questions of the logic of appropriateness: "What kind of situation is this?", "What kind of person is $p_i$?", and "What does a person like $p_i$ do in this situation?" 

To avoid question-order and carryover effects, we reset the agent’s memory after each question. This prevents contextual priming and artificial response consistency, which are known issues in survey studies \citep{schuman1996questions, tourangeau2000psychology}. Finally, the selected responses of each persona are numerically encoded based on the Likert scale, and aggregated by mean along each diversity axis $\mathcal{D}$, representing the persona's response embedding $z_i$.

\vspace{-0.25cm}
\subsection{Diversity Metrics}
\label{diversity_metrics}

Ideally, we want the population vectors $\mathcal{Z}$ to cover the full support of the space spanned by $\mathcal{D}$, including rare or extreme persona configurations. To encourage broad and well-distributed coverage, and to quantify diversity, we define six diversity metrics $\mathcal{M}(\mathcal{Z})$.

Specifically, we aim to maximize: (i) \textbf{coverage}, estimated via Monte Carlo sampling; (ii) \textbf{convex hull volume} the volume of the smallest convex set containing the points $\mathcal{Z}$; (iii) the \textbf{minimum pairwise distance} to ensure that no two personas are identical; and (iv) the \textbf{average pairwise distance} to measure the spread. We also aim to minimize: (v) \textbf{dispersion}, defined as the radius of the largest empty region in the space; and (vi) the \textbf{KL divergence} between the empirical distribution of $\mathcal{Z}$ and an ideal quasi-random reference distribution. These metrics act as complementary geometric proxies that guard against single-metric degenerate solutions rather than exact estimators of psychological support; in particular, convex hull volume does not imply an assumption that the underlying human trait space is convex. More details in Appendix \ref{appendix_diversity_metrics}.

\vspace{-0.25cm}
\subsection{AlphaEvolve as Evolutionary Loop}
\label{alphaevolve_loop}

We use AlphaEvolve \citep{novikov2025alphaevolve}\footnote{Accessible via the \href{https://cloud.google.com/gemini/enterprise/docs/alphaevolve/developer-guide/overview}{Google Cloud API}} to solve the optimization problem defined in Equation (1). We run 10 parallel islands, seeded in a round-robin fashion using the three initial generators. The evolution process runs for 500 iterations. Mutations of the code $\phi$ are performed by Gemini 2.5 Pro \citep{comanici2025gemini}, using the mutation prompts described in Appendix \ref{alphaevolve_details}. AlphaEvolve also sees a detailed system prompt on what the task is about; we show the prompt in Appendix \ref{system_prompt}. We allow the mutation operator to modify the logic within both Stage~1 (autoregressive generation) and Stage~2 (parallel expansion), but not the two-stage structure and their ordering.

Evaluation is performed over a batch of training questionnaires. For each questionnaire $q$, a candidate generator produces a population of $N=25$ personas. We simulate their responses to obtain the embeddings $\mathcal{Z} = \Psi(\mathcal{P}, q)$ and compute the diversity metrics $\mathcal{M}(\mathcal{Z})$ and average them across all questionnaires. Between iterations, the mutation operator is provided with feedback: it observes a random subset of generated persona profiles alongside their corresponding response scores, grounding future code improvements in empirical data.

Although the metrics in $\mathcal{M}$ are correlated, optimizing multiple objectives helps preserve solution diversity. AlphaEvolve maintains separate elites for each metric within each island; with 10 islands and 6 metrics, this results in up to 60 distinct generators being tracked at any point in time. Islands evolve independently with periodic extinction events every 8 hours of wall time (approximately 100 iterations), during which poorly performing islands are reset using solutions copied from the best-performing islands to encourage exploration.


\section{Experimental Setup}

\subsection{Baselines}
We compare the optimized Persona Generator code $\phi^*$ against four baselines: Nemotron Personas \citep{nvidia2025personas}, a Diversity Prompting baseline designed by Claude Opus 4.6 \citep{anthropic2026claude}, a formative memory generator adapted from Concordia \citep{vezhnevets2023generative}, and a name-only baseline.

Nemotron Personas is a static dataset of 100,000 personas grounded in U.S. demographic statistics. Since it is not a generator, we evaluate it by randomly sampling personas from the dataset. The Diversity Prompting baseline was constructed using the Claude Code agentic harness with Claude Opus 4.6 \citep{anthropic2026claude} in high thinking effort to rewrite the same starting two-stage persona generator and diversity objective given to AlphaEvolve. The Concordia baseline constructs personas by generating formative memories from early life onward, with the goal of explaining present behavior through past experiences \citep{vezhnevets2023generative}. Finally, the name-only baseline provides the LLM with only a persona’s name and no additional description, relying on the model’s internal priors to infer an appropriate behavioral profile. All baselines are evaluated using the same Concordia simulation setup and act according to the logic of appropriateness \citep{march2011logic, leibo2024theory} articulated in Section \ref{simulations}. Example persona descriptions are shown in Appendix~\ref{baseline_example}.

\begin{figure}[t]
    \centering
    \begin{subfigure}[b]{0.49\textwidth}
        \centering
        \includegraphics[width=\linewidth]{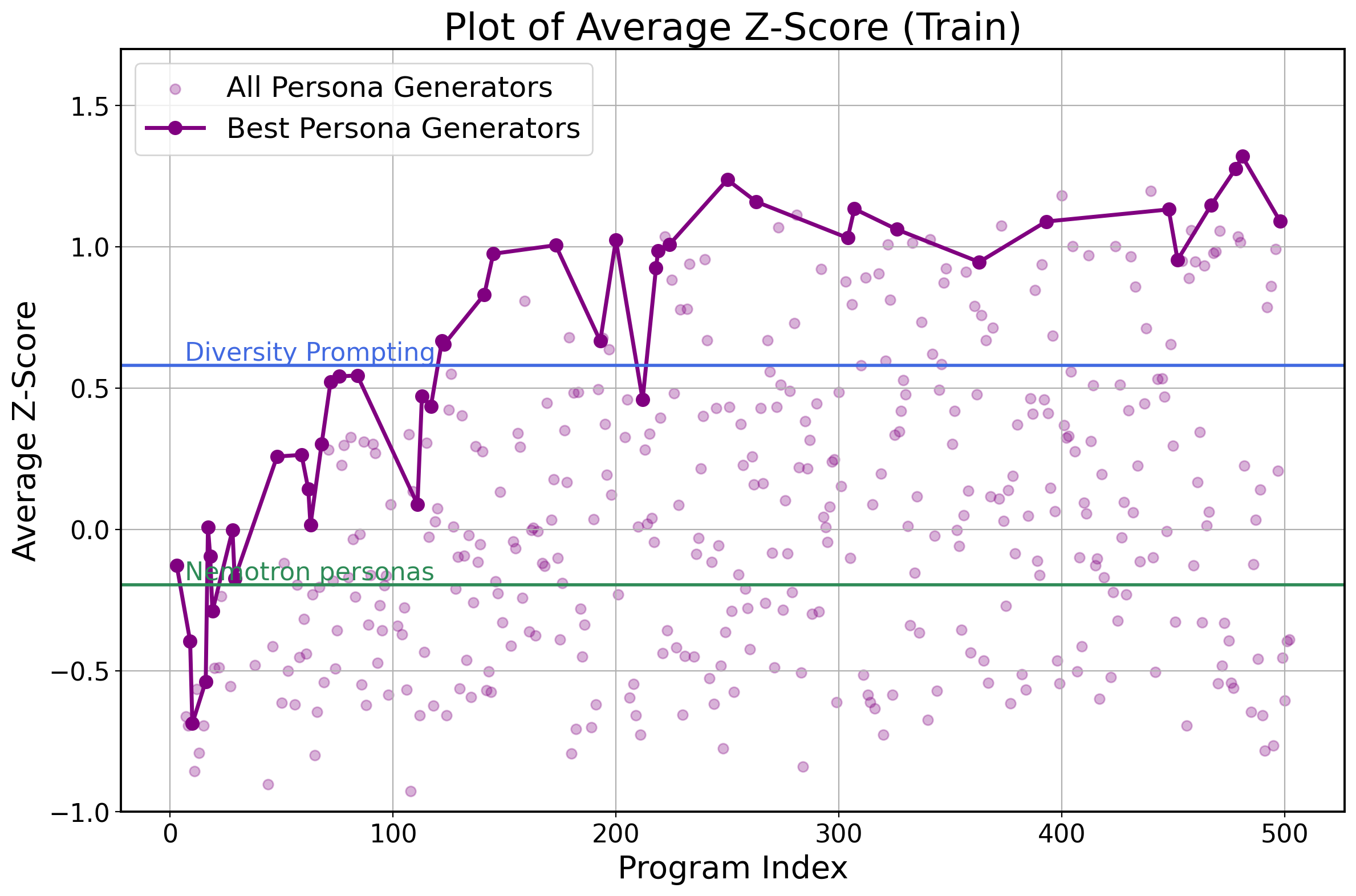} 
        \caption{Performance on Training/Validation sets.}
        \label{fig:train_performance}
    \end{subfigure}
    \hfill
    \begin{subfigure}[b]{0.49\textwidth}
        \centering
        \includegraphics[width=\linewidth]{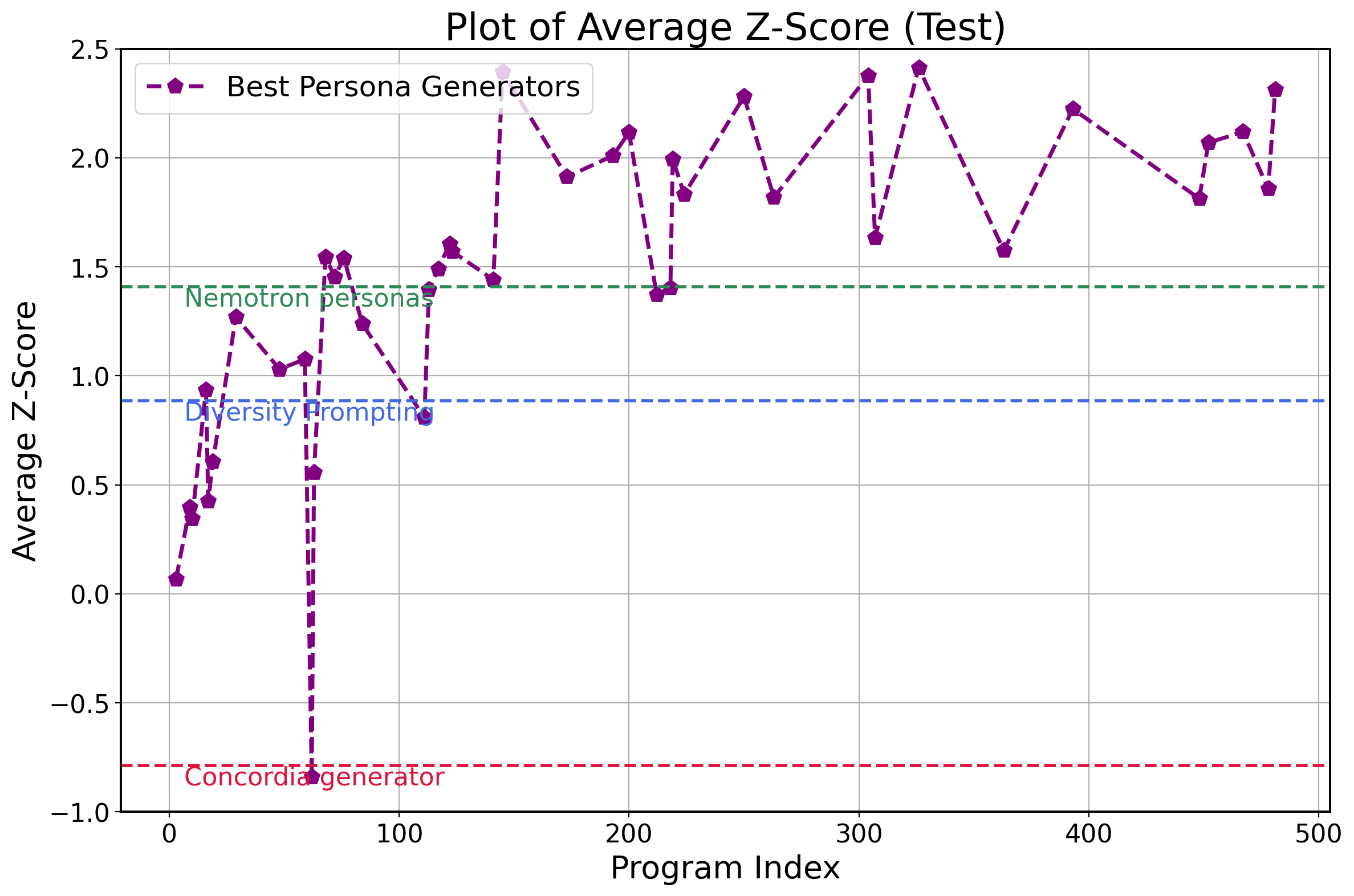} 
        \caption{Generalization on held-out Test set.}
        \label{fig:test_performance}
    \end{subfigure}

    \caption{\textbf{Evolution of Persona Generator performance.} Comparison between training and testing phases. The left panel (a) shows the mean Z-score across the diversity metrics on the 40 training/validation questionnaires, while the right panel (b) displays generalization performance of the best discovered Persona Generators on the 10 held-out test questionnaires.}
    \label{main_figure}
    \vspace{-0.2cm}
\end{figure}
\FloatBarrier

\vspace{-0.2cm}

\subsection{Results}

Figure~\ref{fig:train_performance} shows the results of running AlphaEvolve \citep{novikov2025alphaevolve} on our pipeline. Each point corresponds to an evolved Persona Generator $\phi$, evaluated by generating $N=25$ personas for each of the 40 training/validation questionnaires (1,000 evaluated personas per data point). The solid line tracks the best-performing solutions over time. A generator is marked as a new best whenever it achieves a record score on any of the six diversity metrics. As evolution progresses, AlphaEvolve consistently discovers increasingly effective Persona Generators, leading to substantial improvements in population diversity.

The evolved generators outperform all baselines on every test-set metric, and on every training metric except KL, where Diversity Prompting is marginally better (Table \ref{tab:diversity_results}). The optimized generators also generalize well to held-out test questionnaires (Figure~\ref{fig:test_performance}). Notably, in our implementation, iterative search over generator control flow outperforms hand-engineered generators and a strong single-pass rewrite by a frontier model with an agentic harness (\textit{Diversity Prompting}); we make no claim about AlphaEvolve versus other program optimizers, as the search component is modular. Additional metric plots are provided in Appendix~\ref{results_diversity_metrics}.

\begin{table}[t]
\centering
\caption{Evaluation of Persona Generators and baselines across diversity metrics and aggregate z-score. Results are reported for both training and test sets. \textbf{Bold}\,=\,best, \underline{underline}\,=\,second best.}
\label{tab:diversity_results}
\resizebox{\textwidth}{!}{%
\begin{tabular}{lcccccc}
\toprule
\multicolumn{7}{c}{Train\,/\,Test} \\
Method & Hull & Cov. & Disp. & KL & Dist. & Avg.\ $z$ \\
\midrule
Best Persona Generator & \textbf{0.25}\,/\,\textbf{0.36} & \textbf{0.72}\,/\,\textbf{0.82} & \textbf{-2.80}\,/\,\textbf{-2.68} & \underline{-12.41}\,/\,\textbf{-10.08} & \textbf{2.38}\,/\,\textbf{2.72} & \textbf{1.32}\,/\,\textbf{2.31} \\
Nemotron Personas & 0.12\,/\,\underline{0.24} & 0.56\,/\,\underline{0.73} & -3.72\,/\,\underline{-2.76} & -15.18\,/\,-15.13 & 1.78\,/\,\underline{2.56} & -0.20\,/\,\underline{1.41} \\
Diversity Prompting & \underline{0.15}\,/\,0.18 & \underline{0.68}\,/\,0.65 & \underline{-2.91}\,/\,-2.85 & \textbf{-11.75}\,/\,\underline{-11.86} & \underline{2.03}\,/\,2.41 & \underline{0.58}\,/\,0.89 \\
Concordia Generator & 0.05\,/\,0.05 & 0.45\,/\,0.49 & -3.95\,/\,-3.34 & -22.51\,/\,-14.96 & 1.06\,/\,1.48 & -1.33\,/\,-0.79 \\
Name Baseline & 0.01\,/\,0.02 & 0.37\,/\,0.44 & -4.46\,/\,-3.61 & -36.69\,/\,-24.71 & 0.46\,/\,0.92 & -2.36\,/\,-1.54 \\
\bottomrule
\end{tabular}
}
\vspace{-0.1cm}
\end{table}

\subsection{Persona Realism and Human Fidelity}
\label{sec:realism}
A potential concern when maximizing support coverage is that the generator might produce incoherent, contradictory, or alien personas simply to artificially inflate diversity metrics. To evaluate textual coherence and plausibility, we conducted an LLM-as-a-judge evaluation on 1,000 generated personas from the best evolved solution, using three independent frontier models. As shown in Table \ref{tab:llm_realism}, all three models overwhelmingly judged the personas as realistic for their given contexts (93.0\%--99.2\%). Qualitative inspection of the few flagged cases revealed mildly implausible trait combinations (e.g., a medical student with hedge-fund level finance skills) rather than degenerate or gibberish outputs. Because we evolve the generator code rather than personas directly, the LLM's autoregressive generation and strong pull toward the mean act as an implicit regularizer against incoherent outliers.

Beyond persona coherence, we verify whether this generated diversity aligns with actual human statistical variance. We compared 100 personas per method against 1,000 real human responses on the Big Five Inventory (BFI) \citep{john1999bigfive, open_psychometrics}, never seen during optimization. Using a k-Nearest Neighbors (k-NN) approach calibrated to real-data baselines, we measured the fraction of personas within the human data bounds (\textbf{In Support}), the proportion of the human distribution spanned (\textbf{Coverage}), and the density match (\textbf{KL Divergence}).

As shown in Table \ref{tab:bfi_human_data}, our evolved generator achieves the highest human distribution coverage (62.9\%) and the lowest KL divergence (0.99), while 98\% of its personas remain strictly within realistic human bounds. Crucially, it significantly outperforms baselines explicitly grounded in real-world demographics, such as Nemotron Personas \citep{nvidia2025personas}. Standard LLMs suffer from severe mode collapse, clustering tightly around highly agreeable, stereotypical behaviors. By explicitly optimizing for broad support, our method fights this collapse to populate the long tails that baselines completely miss. Merely inflating variance would push personas outside human support and raise KL; instead, our method increases coverage while lowering KL (and per-trait marginal KL on four of five traits), whereas a 100\% in-support rate can be reached trivially by collapsing to the mean, as the name-only baseline shows. Across 500 bootstrap replicates ($n=80$), our generator beats Diversity Prompting by $+7.0$ coverage points [1.4, 13.0] with $1.20$ [0.94, 1.44] lower KL, and recovers 67.8\% [62.6, 73.1] of the 91.1\% coverage reached by held-out humans scored through the same protocol (ranking invariant across $k \in \{10,\dots,45\}$). This demonstrates that maximizing support coverage is not at odds with density matching; rather, overcoming LLM mode collapse is a fundamental prerequisite for accurately capturing real-world human distributions.
\vspace{-0.3cm}

\begin{table}[ht]
    \centering
    \begin{minipage}[h]{0.42\textwidth}
        \centering
        \caption{\textbf{LLM-as-a-judge realism.} Evaluation of 1,000 generated personas. The vast majority were deemed realistic across three frontier models, confirming our objective does not yield degenerate edge cases.}
        \label{tab:llm_realism}
        \vspace{0.1cm} 
        \small 
        \begin{tabular}{@{}lcc@{}}
            \toprule
            \textbf{Judge Model} & \textbf{Realistic} & \textbf{Implausible} \\
            \midrule
            Gemini 3 Flash & 99.2\% & 0.8\% \\
            Claude 4.5 Haiku & 97.7\% & 2.3\% \\
            GPT 5.4 Mini & 93.0\% & 7.0\% \\
            \bottomrule
        \end{tabular}
    \end{minipage}\hfill
    \begin{minipage}[h]{0.55\textwidth}
        \centering
        \caption{\textbf{Comparison to real human BFI data.} The best evolved persona generator generates personas closer to the true density of human behavior (lowest KL) without generating implausible outliers (98\% in-support).}
        \label{tab:bfi_human_data}
        \vspace{0.1cm} 
        \small 
        \begin{tabular}{@{}lccc@{}}
            \toprule
            \textbf{Method} & \textbf{In Support} $\uparrow$ & \textbf{Coverage} $\uparrow$ & \textbf{KL} $\downarrow$ \\
            \midrule
            \textbf{Persona Generator} & 98.0\% & \textbf{62.9\%} & \textbf{0.99} \\
            Diversity Prompting & 100.0\% & 54.1\% & 2.16 \\
            Nemotron Personas & 100.0\% & 51.5\% & 3.00 \\
            Concordia Generator & 100.0\% & 44.0\% & 3.73 \\
            Name Baseline & 100.0\% & 28.1\% & 5.03 \\
            \bottomrule
        \end{tabular}
    \end{minipage}
    \vspace{-0.2cm}
\end{table}

\subsection{Transfer to Manifested Behavior}
\label{downstream_tasks_section}
While questionnaires robustly measure stated preferences, we also evaluate whether diversity gains transfer to manifested behavior in open-ended text. Measuring semantic diversity in unconstrained text is challenging, but qualitative differences in behavioral mode-collapse are stark. We evaluated a comedy setting (personas telling jokes) and a stressful conflict resolution scenario (a car crash).

When embedded using the Gemini Embedder 2 \citep{lee2025gemini}, evolved generators exhibit broader metric coverage across both tasks compared to baselines (see Appendix \ref{downstream_tasks_appendix}). Importantly, qualitative analysis (detailed in Appendices \ref{umap_appendix} and \ref{downstream_tasks_qualitative_appendix}, including the comedy UMAP projection in Figure~\ref{umap_figure}) reveals that baselines suffer from severe behavioral mode collapse. In conflict resolution, baseline personas default to being highly conflict-averse, polite, and collaborative, even when their car is badly damaged by a distracted driver. Conversely, evolved generators produce a realistic spectrum of emotions, including genuine verbal frustration and varied coping mechanisms. Similarly, baseline comedy outputs cluster around a few repetitive joke formats, whereas evolved generators span dry, dark, and witty humor. This confirms that structural diversity optimization successfully mitigates the narrow behavioral guardrails typical of standard LLM outputs.

To further test non-geometric downstream utility, we run two open-ended product stress tests using Gemini 3.5 Flash for both generation and simulation ($N=25$ personas over 10 seeds, testing zero-shot cross-domain and cross-model transfer simultaneously): \textbf{PlateMate}, a food-delivery app billing \$140/week for a 7-day meal plan with fixed delivery windows, and \textbf{MoveMate}, an AI relocation assistant proposing a default London move plan (£1,800/month neighborhood, visa checklist, banking/telecom setup, shipping, and timeline). Personas react in character to silent assumptions in the default plans; an LLM judge extracts friction points, which are clustered into a single pooled taxonomy across all methods. As shown in Table~\ref{tab:stress_tests}, the evolved Persona Generator surfaces the largest share of the failure taxonomy on both tasks (within error of Diversity Prompting on PlateMate; improving over its Concordia seed from 61.7\% to 76.9\% on PlateMate and 68.9\% to 78.9\% on MoveMate) while maintaining high plausibility (4.68--4.85/5). Notably, Nemotron Personas drops sharply on MoveMate (70.5\% to 53.5\%), reflecting poor transfer of static U.S. demographics to a U.K. scenario, while the Name Baseline achieves the highest plausibility (4.96) with only 15.7\% coverage, showing that bland responses trivially maximize plausibility. Qualitatively, our populations surfaced edge-case failures missed by defaults, such as shared-fridge limits in flatshares and seniors needing landline ordering on PlateMate, or visa timelines colliding with school enrollment and deposits due before first salary on MoveMate.

\begin{table}[ht]
\centering
\caption{\textbf{Downstream Product Stress Tests.} Fraction of pooled failure taxonomy surfaced and response plausibility (1--5 scale) across 10 seeds ($N=25$, mean $\pm$ s.e.m.) using Gemini 3.5 Flash.}
\label{tab:stress_tests}
\small
\begin{tabular}{@{}lcccc@{}}
\toprule
\textbf{Method} & \textbf{PlateMate Cov.} $\uparrow$ & \textbf{PlateMate Plaus.} & \textbf{MoveMate Cov.} $\uparrow$ & \textbf{MoveMate Plaus.} \\
\midrule
\textbf{Persona Generator} & \textbf{76.9\% $\pm$ 2.4\%} & 4.68 $\pm$ 0.03 & \textbf{78.9\% $\pm$ 2.1\%} & 4.85 $\pm$ 0.01 \\
Diversity Prompting & \underline{73.8\% $\pm$ 2.4\%} & 4.64 $\pm$ 0.03 & \underline{73.7\% $\pm$ 1.8\%} & 4.90 $\pm$ 0.01 \\
Nemotron Personas & 70.5\% $\pm$ 2.7\% & 4.71 $\pm$ 0.03 & 53.5\% $\pm$ 2.2\% & 4.82 $\pm$ 0.01 \\
Concordia Generator & 61.7\% $\pm$ 2.1\% & 4.65 $\pm$ 0.03 & 68.9\% $\pm$ 1.7\% & 4.83 $\pm$ 0.02 \\
Name Baseline & 15.7\% $\pm$ 2.0\% & \textbf{4.96 $\pm$ 0.01} & 46.9\% $\pm$ 1.8\% & 4.82 $\pm$ 0.01 \\
\bottomrule
\end{tabular}
\vspace{-0.2cm}
\end{table}

\subsection{Analysis of Evolved Programs}
\label{sec:analysis}
Across independent AlphaEvolve islands, evolution repeatedly converged on four structural changes, present in most top-performing generators relative to the initial seeds: (i) shifting persona descriptions from third-person to first-person narrative; (ii) adopting the Quasi-Random Monte Carlo sampling scheme from the third seed as the default Stage~1 foundation; (iii) replacing early-life formative memories with explicit descriptions of belief systems and everyday decision heuristics; and (iv) grounding personas with concrete examples of how they behave across scenarios.
The top solution by average overall score produces first-person personas, written as short paragraphs that describe a persona’s internal reasoning across situations. This differs from the basic Concordia agents used in our simulations, which rely on third-person role-playing \citep{vezhnevets2023generative}. The top solution by convex hull volume also uses a first-person perspective, expressed in a more rule-based format. By contrast, the top solution by average coverage produces third-person descriptions that focus on a persona’s core motivations and their logic of appropriateness. We provide examples in Appendix \ref{evolved_solutions} (and baseline examples in Appendix~\ref{baseline_example}) and release the top-performing generators.\footnote{\url{https://github.com/google-deepmind/concordia/tree/main/concordia/contrib/persona_generators}}

Importantly, the Persona Generators discovered by AlphaEvolve do not overfit to the optimization model. As shown in Table \ref{tab:cross_model}, when we substitute the underlying Gemma 3 27B inference model with Gemini 3.1 Flash-Lite or Qwen 3.5 27B \citep{qwen35blog}, the evolved code maintains its high coverage and diversity margins. In fact, the transferred models even exceed the original model on metrics like Hull Volume and Mean Distance, demonstrating that the learned generation logic is structurally robust and transfers effectively across entirely different model families.
\vspace{-0.3cm}

\begin{table}[ht]
\centering
\caption{\textbf{Cross-Model Transfer.} Evaluation of the best evolved generator code (optimized on Gemma) across different inference models on the test set.}
\label{tab:cross_model}
\small 
\begin{tabular}{@{}lccccc@{}}
\toprule
\textbf{Inference Model} & \textbf{Hull Vol.}$\uparrow$ & \textbf{Cov.}$\uparrow$ & \textbf{Disp.}$\downarrow$ & \textbf{KL}$\downarrow$ & \textbf{Dist.}$\uparrow$ \\
\midrule
Best Persona Generator (Gemma 3 27B) & 0.36 & \textbf{0.82} & -2.68 & \textbf{-10.08} & 2.72 \\
Best Persona Generator (Gemini 3.1 Flash-Lite) & 0.47 & 0.75 & \textbf{-2.75} & -18.63 & 3.14 \\
Best Persona Generator (Qwen 3.5 27B) & \textbf{0.61} & 0.81 & -2.52 & -12.88 & \textbf{3.30} \\
\bottomrule
\end{tabular}
\vspace{-0.2cm} 
\end{table}

The full evolutionary search required approximately 1,000 Gemini 2.5 Pro API calls (for mutations) and 25M Gemma 3 27B calls (for persona generation and simulation; $\sim$1--2K input tokens each to an open-weights 27B model), taking roughly 3 days of wall-clock time on a TPU v4 node \citep{jouppi2023tpu}. Crucially, this is a one-time optimization cost. The resulting generator functions are lightweight and generate new populations at inference time with minimal compute.


\section{Discussion}

\paragraph{Measuring Unconstrained Behavioral Diversity}
While automated questionnaires provide a robust, quantifiable testbed for optimization, evaluating behavioral diversity in unconstrained, open-ended text remains a fundamental challenge for the field. Standard embedding-based metrics are notoriously noisy in these settings, as they struggle to disentangle meaningful semantic diversity from mere stylistic variance or incoherence. However, as shown in Section \ref{downstream_tasks_section}, our evolved generators clearly break the polite, conflict-averse mode-collapse typical of standard LLMs, producing a much wider spectrum of realistic reactions. Developing rigorous, noise-free metrics for open-ended text diversity, alongside direct human evaluations of perceived realism and diversity across additional psychometric instruments, is a crucial next step for the community. Such metrics would unlock the ability to optimize directly for diversity on complex downstream environments, rather than using stated preferences as an optimization proxy.

\paragraph{Decoupling Optimization from Inference Scale}
By design, we decoupled the optimization phase from the deployment (inference) phase. Running AlphaEvolve with $N=25$ personas across 50 distinct contexts allowed us to prioritize broad zero-shot generalization over deep specialization in a single domain. Importantly, the optimization costs do not cap the inference, because our output is a reusable Python function, researchers can trivially scale to thousands of personas by running the generator multiple times in a loop. Furthermore, as shown in Appendix \ref{scaling_100}, our generators successfully scale their zero-shot batches to $N=100$ while maintaining diversity margins over baselines. An exciting future work direction is extreme-scale one-shot generation (e.g., $N\ge10,000$) in a single forward pass.

\paragraph{Influence of Mutation Prompts}
Finally, while AlphaEvolve discovered highly effective generation algorithms, the evolutionary search space is inherently biased by the LLM mutation operator's internal priors. While this allowed for rapid discovery of efficient generator code, more open-ended meta-learning approaches that dynamically adapt the mutation strategies could yield even more unconventional and diverse generation algorithms in the future.


\section{Related Work}

\paragraph{Generative Agent-Based Modeling and Algorithmic Fidelity}
Generative agent-based modeling (GABM) has emerged as a powerful paradigm for simulating complex social interactions. Its validity relies on \emph{algorithmic fidelity}, a term describing how accurately an LLM can reproduce the beliefs, attitudes, and response patterns of specific human sub-populations \citep{argyle2023out}. By conditioning LLMs on demographic backstories \citep{argyle2023out} or interview transcripts \citep{park2024generative}, agents can store memories, reflect \citep{park2023generative}, and accurately mirror human behaviors in both psychometric surveys and interactive economic games \citep{slumbers2025using}. This approach has been scaled to datasets of 100,000 \citep{nvidia2025personas} and 1B personas \citep{ge2024scaling}, with further work improving narrative depth via human-attribute taxonomies \citep{wang2025deeppersona}. To manage the architectural complexity of these multi-agent systems, \citet{vezhnevets2023generative} developed Concordia \citep{vezhnevets2025multi}, while \citet{leibo2024theory} operationalize the ``logic of appropriateness'' \citep{march2011logic} to configure agent actions. Beyond prompting, steering vector techniques offer an alternative way to shape behaviors by directly modulating LLMs' hidden activations \citep{turner2023steering, rimsky2024steering, templeton2024scaling, zou2023representation, chen2025persona}. However, synthetic personas remain prone to systematic biases \citep{li2025llm, venkit2025tale}: they can fail to align with the opinions of some demographic groups even when explicitly steered \citep{santurkar2023whose}, and can struggle to match the authentic style of human speech \citep{amirova2024framework}.

\paragraph{Measuring Preferences and Psychometric Tests}
Likert-style questionnaires and established inventories, like the Big Five, are standard tools for measuring human traits and preferences \citep{likert1932technique, nunnally1978overview, john1999bigfive, murphy2011measuring}. In agent-based modeling, they are widely used to evaluate synthetic personas, which can complete these surveys while exhibiting consistent profiles \citep{jiang2024personallm, bhandari2025evaluating}. Psychometric tools benchmark algorithmic fidelity by comparing simulated personas against real individuals \citep{park2024generative}, test role-play stability \citep{wang2025evaluating}, and validate the behavioral coherence of synthetic populations \citep{pellert2024ai, maharjan2025psychometric, castricato2025persona, bisbee2024synthetic, petrov2024limited}. Finally, recent work explores using LLMs to automatically generate such questionnaires \citep{adhikari2025exploring}.

\paragraph{Evolution and Discovery with LLMs}
LLMs have proven to be strong evolutionary operators across various domains. They can effectively mutate and refine text in self-referential loops \citep{fernando2023promptbreeder} and drive quality-diversity search in creative tasks \citep{bradley2023quality}. Prompt evolution is also widely applied to AI safety and red-teaming to discover adversarial prompts and jailbreaks \citep{samvelyan2024rainbow, wang2025quality}, sometimes conditioning mutations on simulated user profiles \citep{deng2025personateaming}. Beyond text, LLMs can iteratively generate, recombine, and improve executable code and algorithms \citep{romera2024mathematical}. Frameworks like AlphaEvolve scale this to large-population evolutionary search to discover algorithms and optimize real-world systems \citep{novikov2025alphaevolve}, while similar loops can even automate end-to-end scientific research pipelines \citep{lu2024ai}. Together, these works establish LLMs as powerful mutation engines capable of discovering and improving novel solutions across text, code, and algorithms.

\vspace{-0.1cm}
\section{Conclusion}
We introduced Persona Generators, functions that produce diverse synthetic populations tailored to arbitrary contexts and optimized through evolutionary search. Evolved generators substantially outperform existing baselines across six diversity metrics and generalize to held-out contexts, producing populations that span rare trait combinations difficult to elicit through standard prompting. Our results show that optimizing the generator itself, rather than individual personas or fixed populations, is a viable and promising direction. The resulting Persona Generators are lightweight and reusable, enabling on-demand diverse population synthesis. We open-source the top-performing implementations to support further research.

Finally, while these tools can be used for stress-testing AI, red-teaming, and generating diverse interactive agents for virtual environments, they also carry dual-use risks; we provide a discussion of broader societal impacts in Appendix \ref{impact_statement}.

\begin{ack}
We thank George Tsoukalas, Emilien Dupont, Marvin Eisenberger, Alexander Novikov, and Po-Sen Huang from the AlphaEvolve team, as well as Michael Dennis, for the insightful discussions.
\end{ack}

\bibliography{bibliography}

@article{vezhnevets2023generative,
  title={Generative agent-based modeling with actions grounded in physical, social, or digital space using Concordia},
  author={Vezhnevets, Alexander Sasha and Agapiou, John P and Aharon, Avia and Ziv, Ron and Matyas, Jayd and Du{\'e}{\~n}ez-Guzm{\'a}n, Edgar A and Cunningham, William A and Osindero, Simon and Karmon, Danny and Leibo, Joel Z},
  journal={arXiv preprint arXiv:2312.03664},
  year={2023}
}

@article{chen2025persona,
  title={Persona vectors: Monitoring and controlling character traits in language models},
  author={Chen, Runjin and Arditi, Andy and Sleight, Henry and Evans, Owain and Lindsey, Jack},
  journal={arXiv preprint arXiv:2507.21509},
  year={2025}
}

@article{park2024generative,
  title={Generative agent simulations of 1,000 people},
  author={Park, Joon Sung and Zou, Carolyn Q and Shaw, Aaron and Hill, Benjamin Mako and Cai, Carrie and Morris, Meredith Ringel and Willer, Robb and Liang, Percy and Bernstein, Michael S},
  journal={arXiv preprint arXiv:2411.10109},
  year={2024}
}

@inproceedings{park2023generative,
  title={Generative agents: Interactive simulacra of human behavior},
  author={Park, Joon Sung and O'Brien, Joseph and Cai, Carrie Jun and Morris, Meredith Ringel and Liang, Percy and Bernstein, Michael S},
  booktitle={Proceedings of the 36th annual acm symposium on user interface software and technology},
  pages={1--22},
  year={2023}
}

@article{fernando2023promptbreeder,
  title={Promptbreeder: Self-referential self-improvement via prompt evolution},
  author={Fernando, Chrisantha and Banarse, Dylan and Michalewski, Henryk and Osindero, Simon and Rockt{\"a}schel, Tim},
  journal={arXiv preprint arXiv:2309.16797},
  year={2023}
}

@article{bradley2023quality,
  title={Quality-diversity through AI feedback},
  author={Bradley, Herbie and Dai, Andrew and Teufel, Hannah and Zhang, Jenny and Oostermeijer, Koen and Bellagente, Marco and Clune, Jeff and Stanley, Kenneth and Schott, Gr{\'e}gory and Lehman, Joel},
  journal={arXiv preprint arXiv:2310.13032},
  year={2023}
}

@article{samvelyan2024rainbow,
  title={Rainbow teaming: Open-ended generation of diverse adversarial prompts},
  author={Samvelyan, Mikayel and Raparthy, Sharath C and Lupu, Andrei and Hambro, Eric and Markosyan, Aram H and Bhatt, Manish and Mao, Yuning and Jiang, Minqi and Parker-Holder, Jack and Foerster, Jakob and others},
  journal={Advances in Neural Information Processing Systems},
  volume={37},
  pages={69747--69786},
  year={2024}
}

@article{novikov2025alphaevolve,
  title={AlphaEvolve: A coding agent for scientific and algorithmic discovery},
  author={Novikov, Alexander and V{\~u}, Ng{\^a}n and Eisenberger, Marvin and Dupont, Emilien and Huang, Po-Sen and Wagner, Adam Zsolt and Shirobokov, Sergey and Kozlovskii, Borislav and Ruiz, Francisco JR and Mehrabian, Abbas and others},
  journal={arXiv preprint arXiv:2506.13131},
  year={2025}
}

@article{romera2024mathematical,
  title={Mathematical discoveries from program search with large language models},
  author={Romera-Paredes, Bernardino and Barekatain, Mohammadamin and Novikov, Alexander and Balog, Matej and Kumar, M Pawan and Dupont, Emilien and Ruiz, Francisco JR and Ellenberg, Jordan S and Wang, Pengming and Fawzi, Omar and others},
  journal={Nature},
  volume={625},
  number={7995},
  pages={468--475},
  year={2024},
  publisher={Nature Publishing Group UK London}
}

@article{lu2024ai,
  title={The ai scientist: Towards fully automated open-ended scientific discovery},
  author={Lu, Chris and Lu, Cong and Lange, Robert Tjarko and Foerster, Jakob and Clune, Jeff and Ha, David},
  journal={arXiv preprint arXiv:2408.06292},
  year={2024}
}

@software{nvidia2025personas,
  author = {Meyer, Yev and Corneil, Dane},
  title = {{Nemotron-Personas-USA}: Synthetic Personas Aligned to Real-World Distributions
},
  month = {June},
  year = {2025},
  url = {https://huggingface.co/datasets/nvidia/Nemotron-Personas-USA}
}

@article{vezhnevets2025multi,
  title={Multi-Actor Generative Artificial Intelligence as a Game Engine},
  author={Vezhnevets, Alexander Sasha and Matyas, Jayd and Cross, Logan and Paglieri, Davide and Chang, Minsuk and Cunningham, William A and Osindero, Simon and Isaac, William S and Leibo, Joel Z},
  journal={arXiv preprint arXiv:2507.08892},
  year={2025}
}

@article{li2025llm,
  title={LLM Generated Persona is a Promise with a Catch},
  author={Li, Ang and Chen, Haozhe and Namkoong, Hongseok and Peng, Tianyi},
  journal={arXiv preprint arXiv:2503.16527},
  year={2025}
}

@article{leibo2024theory,
  title={A theory of appropriateness with applications to generative artificial intelligence},
  author={Leibo, Joel Z and Vezhnevets, Alexander Sasha and Diaz, Manfred and Agapiou, John P and Cunningham, William A and Sunehag, Peter and Haas, Julia and Koster, Raphael and Du{\'e}{\~n}ez-Guzm{\'a}n, Edgar A and Isaac, William S and others},
  journal={arXiv preprint arXiv:2412.19010},
  year={2024}
}

@incollection{march2011logic,
    author = {March, James G. and Olsen, Johan P.},
    title = "{The Logic of Appropriateness}",
    booktitle = "{The Oxford Handbook of Political Science}",
    publisher = {Oxford University Press},
    year = {2011},
    doi = {10.1093/oxfordhb/9780199604456.013.0024},
}

@article{turner2023steering,
  title={Steering language models with activation engineering},
  author={Turner, Alexander Matt and Thiergart, Lisa and Leech, Gavin and Udell, David and Vazquez, Juan J and Mini, Ulisse and MacDiarmid, Monte},
  journal={arXiv preprint arXiv:2308.10248},
  year={2023}
}

@inproceedings{rimsky2024steering,
  title={Steering llama 2 via contrastive activation addition},
  author={Rimsky, Nina and Gabrieli, Nick and Schulz, Julian and Tong, Meg and Hubinger, Evan and Turner, Alexander},
  booktitle={Proceedings of the 62nd Annual Meeting of the Association for Computational Linguistics (Volume 1: Long Papers)},
  pages={15504--15522},
  year={2024}
}

@inproceedings{cross2026generative_cogsci,
  title={A Generative Model of Conspicuous Consumption and Status Signaling},
  author={Cross, Logan and Grau-Moya, Jordi and Cunningham, William A. and Vezhnevets, Alexander Sasha and Leibo, Joel Z.},
  booktitle={Proceedings of the 48th Annual Conference of the Cognitive Science Society},
  year={2026}
}

@article{matyas2026stabilising,
  title={Stabilising Generative Models of Attitude Change},
  author={Matyas, Jayd and Cunningham, William A. and Vezhnevets, Alexander Sasha and Mobbs, Dean and Du{\'e}{\~n}ez-Guzm{\'a}n, Edgar A. and Leibo, Joel Z.},
  journal={arXiv preprint arXiv:2604.19791},
  year={2026},
  url={https://arxiv.org/abs/2604.19791}
}

@article{kozlowski2024insilico,
  title={In silico sociology: forecasting COVID-19 polarization with large language models},
  author={Kozlowski, Austin C. and Kwon, Hyunku and Evans, James A.},
  journal={arXiv preprint arXiv:2407.11190},
  year={2024},
  url={https://arxiv.org/abs/2407.11190}
}

@book{templeton2024scaling,
  title={Scaling monosemanticity: Extracting interpretable features from claude 3 sonnet},
  author={Templeton, Adly},
  year={2024},
  publisher={Anthropic}
}

@article{zou2023representation,
  title={Representation engineering: A top-down approach to ai transparency},
  author={Zou, Andy and Phan, Long and Chen, Sarah and Campbell, James and Guo, Phillip and Ren, Richard and Pan, Alexander and Yin, Xuwang and Mazeika, Mantas and Dombrowski, Ann-Kathrin and others},
  journal={arXiv preprint arXiv:2310.01405},
  year={2023}
}

@article{argyle2023out,
  title={Out of one, many: Using language models to simulate human samples},
  author={Argyle, Lisa P and Busby, Ethan C and Fulda, Nancy and Gubler, Joshua R and Rytting, Christopher and Wingate, David},
  journal={Political Analysis},
  volume={31},
  number={3},
  pages={337--351},
  year={2023},
  publisher={Cambridge University Press}
}

@article{amirova2024framework,
  title={Framework-based qualitative analysis of free responses of Large Language Models: Algorithmic fidelity},
  author={Amirova, Aliya and Fteropoulli, Theodora and Ahmed, Nafiso and Cowie, Martin R and Leibo, Joel Z},
  journal={Plos one},
  volume={19},
  number={3},
  pages={e0300024},
  year={2024},
  publisher={Public Library of Science San Francisco, CA USA}
}

@article{slumbers2025using,
  title={Using Large Language Models to Simulate Human Behavioural Experiments: Port of Mars},
  author={Slumbers, Oliver and Leibo, Joel Z and Janssen, Marco A},
  journal={arXiv preprint arXiv:2506.05555},
  year={2025}
}

@inproceedings{santurkar2023whose,
  title={Whose opinions do language models reflect?},
  author={Santurkar, Shibani and Durmus, Esin and Ladhak, Faisal and Lee, Cinoo and Liang, Percy and Hashimoto, Tatsunori},
  booktitle={International Conference on Machine Learning},
  pages={29971--30004},
  year={2023},
  organization={PMLR}
}

@article{likert1932technique,
  author = {Likert, Rensis},
  title = {{A technique for the measurement of attitudes}},
  journal = {Archives of Psychology},
  volume = {22},
  number = {140},
  pages = {5--55},
  year = {1932}
}

@incollection{john1999bigfive,
  author = {John, Oliver P. and Srivastava, Sanjay},
  title = {The Big Five Trait taxonomy: History, measurement, and theoretical perspectives},
  booktitle = {Handbook of Personality: Theory and Research},
  editor = {Pervin, Lawrence A. and John, Oliver P.},
  edition = {2nd},
  pages = {102--138},
  publisher = {Guilford Press},
  address = {New York},
  year = {1999}
}

@article{nunnally1978overview,
  title={An overview of psychological measurement},
  author={Nunnally, Jum C},
  journal={Clinical diagnosis of mental disorders: A handbook},
  pages={97--146},
  year={1978},
  publisher={Springer}
}

@inproceedings{jiang2024personallm,
  title={PersonaLLM: Investigating the ability of large language models to express personality traits},
  author={Jiang, Hang and Zhang, Xiajie and Cao, Xubo and Breazeal, Cynthia and Roy, Deb and Kabbara, Jad},
  booktitle={Findings of the association for computational linguistics: NAACL 2024},
  pages={3605--3627},
  year={2024}
}

@inproceedings{bhandari2025evaluating,
  title={Evaluating personality traits in large language models: Insights from psychological questionnaires},
  author={Bhandari, Pranav and Naseem, Usman and Datta, Amitava and Fay, Nicolas and Nasim, Mehwish},
  booktitle={Companion Proceedings of the ACM on Web Conference 2025},
  pages={868--872},
  year={2025}
}

@article{pellert2024ai,
  title={Ai psychometrics: Assessing the psychological profiles of large language models through psychometric inventories},
  author={Pellert, Max and Lechner, Clemens M and Wagner, Claudia and Rammstedt, Beatrice and Strohmaier, Markus},
  journal={Perspectives on Psychological Science},
  volume={19},
  number={5},
  pages={808--826},
  year={2024},
  publisher={Sage Publications Sage CA: Los Angeles, CA}
}

@article{maharjan2025psychometric,
  title={Psychometric evaluation of large language model embeddings for personality trait prediction},
  author={Maharjan, Julina and Jin, Ruoming and Zhu, Jianfeng and Kenne, Deric},
  journal={Journal of Medical Internet Research},
  volume={27},
  pages={e75347},
  year={2025},
  publisher={JMIR Publications Toronto, Canada}
}

@inproceedings{adhikari2025exploring,
  title={Exploring llms for automated generation and adaptation of questionnaires},
  author={Adhikari, Divya Mani and Hartland, Alexander and Weber, Ingmar and Cannanure, Vikram Kamath},
  booktitle={Proceedings of the 7th ACM Conference on Conversational User Interfaces},
  pages={1--23},
  year={2025}
}

@article{comanici2025gemini,
  title={Gemini 2.5: Pushing the frontier with advanced reasoning, multimodality, long context, and next generation agentic capabilities},
  author={Comanici, Gheorghe and Bieber, Eric and Schaekermann, Mike and Pasupat, Ice and Sachdeva, Noveen and Dhillon, Inderjit and Blistein, Marcel and Ram, Ori and Zhang, Dan and Rosen, Evan and others},
  journal={arXiv preprint arXiv:2507.06261},
  year={2025}
}

@article{messick1968motivational,
  title={Motivational bases of choice in experimental games},
  author={Messick, David M and McClintock, Charles G},
  journal={Journal of experimental social psychology},
  volume={4},
  number={1},
  pages={1--25},
  year={1968},
  publisher={Elsevier}
}

@article{murphy2011measuring,
  title={Measuring social value orientation},
  author={Murphy, Ryan O and Ackermann, Kurt A and Handgraaf, Michel JJ},
  journal={Judgment and Decision making},
  volume={6},
  number={8},
  pages={771--781},
  year={2011},
  publisher={Cambridge University Press}
}

@article{lovibond1995structure,
  title={The structure of negative emotional states: Comparison of the Depression Anxiety Stress Scales (DASS) with the Beck Depression and Anxiety Inventories},
  author={Lovibond, Peter F and Lovibond, Sydney H},
  journal={Behaviour research and therapy},
  volume={33},
  number={3},
  pages={335--343},
  year={1995},
  publisher={Elsevier}
}

@article{webster1994individual,
  title={Individual differences in need for cognitive closure.},
  author={Webster, Donna M and Kruglanski, Arie W},
  journal={Journal of personality and social psychology},
  volume={67},
  number={6},
  pages={1049},
  year={1994},
  publisher={American Psychological Association}
}

@article{team2025gemma,
  title={Gemma 3 technical report},
  author={Team, Gemma and Kamath, Aishwarya and Ferret, Johan and Pathak, Shreya and Vieillard, Nino and Merhej, Ramona and Perrin, Sarah and Matejovicova, Tatiana and Ram{\'e}, Alexandre and Rivi{\`e}re, Morgane and others},
  journal={arXiv preprint arXiv:2503.19786},
  year={2025}
}

@article{wang2025evaluating,
  title={Evaluating the ability of large language models to emulate personality},
  author={Wang, Yilei and Zhao, Jiabao and Ones, Deniz S and He, Liang and Xu, Xin},
  journal={Scientific reports},
  volume={15},
  number={1},
  pages={519},
  year={2025},
  publisher={Nature Publishing Group UK London}
}

@book{schuman1996questions,
  title={Questions and answers in attitude surveys: Experiments on question form, wording, and context},
  author={Schuman, Howard and Presser, Stanley},
  year={1996},
  publisher={Sage}
}

@article{tourangeau2000psychology,
  title={The psychology of survey response},
  author={Tourangeau, Roger and Rips, Lance J and Rasinski, Kenneth},
  year={2000},
  publisher={Cambridge University Press}
}

@article{lee2025gemini,
  title={Gemini embedding: Generalizable embeddings from gemini},
  author={Lee, Jinhyuk and Chen, Feiyang and Dua, Sahil and Cer, Daniel and Shanbhogue, Madhuri and Naim, Iftekhar and {\'A}brego, Gustavo Hern{\'a}ndez and Li, Zhe and Chen, Kaifeng and Vera, Henrique Schechter and others},
  journal={arXiv preprint arXiv:2503.07891},
  year={2025}
}

@article{ge2024scaling,
  title={Scaling synthetic data creation with 1,000,000,000 personas},
  author={Ge, Tao and Chan, Xin and Wang, Xiaoyang and Yu, Dian and Mi, Haitao and Yu, Dong},
  journal={arXiv preprint arXiv:2406.20094},
  year={2024}
}

@article{wang2025deeppersona,
  title={DeepPersona: A Generative Engine for Scaling Deep Synthetic Personas},
  author={Wang, Zhen and Zhou, Yufan and Luo, Zhongyan and Ye, Lyumanshan and Wood, Adam and Yao, Man and Pan, Luoshang},
  journal={arXiv preprint arXiv:2511.07338},
  year={2025}
}

@article{venkit2025tale,
  title={A Tale of Two Identities: An Ethical Audit of Human and AI-Crafted Personas},
  author={Venkit, Pranav Narayanan and Li, Jiayi and Zhou, Yingfan and Rajtmajer, Sarah and Wilson, Shomir},
  journal={arXiv preprint arXiv:2505.07850},
  year={2025}
}

@inproceedings{castricato2025persona,
  title={Persona: A reproducible testbed for pluralistic alignment},
  author={Castricato, Louis and Lile, Nathan and Rafailov, Rafael and Fr{\"a}nken, Jan-Philipp and Finn, Chelsea},
  booktitle={Proceedings of the 31st International Conference on Computational Linguistics},
  pages={11348--11368},
  year={2025}
}

@article{anthis2025llm,
  title={Llm social simulations are a promising research method},
  author={Anthis, Jacy Reese and Liu, Ryan and Richardson, Sean M and Kozlowski, Austin C and Koch, Bernard and Evans, James and Brynjolfsson, Erik and Bernstein, Michael},
  journal={arXiv preprint arXiv:2504.02234},
  year={2025}
}

@article{bisbee2024synthetic,
  title={Synthetic replacements for human survey data? the perils of large language models},
  author={Bisbee, James and Clinton, Joshua D and Dorff, Cassy and Kenkel, Brenton and Larson, Jennifer M},
  journal={Political Analysis},
  volume={32},
  number={4},
  pages={401--416},
  year={2024},
  publisher={Cambridge University Press}
}

@article{petrov2024limited,
  title={Limited ability of llms to simulate human psychological behaviours: a psychometric analysis},
  author={Petrov, Nikolay B and Serapio-Garc{\'\i}a, Gregory and Rentfrow, Jason},
  journal={arXiv preprint arXiv:2405.07248},
  year={2024}
}

@article{deng2025personateaming,
  title={Personateaming: Exploring how introducing personas can improve automated ai red-teaming},
  author={Deng, Wesley Hanwen and Kim, Sunnie SY and Jha, Akshita and Holstein, Ken and Eslami, Motahhare and Wilcox, Lauren and Gatys, Leon A},
  journal={arXiv preprint arXiv:2509.03728},
  year={2025}
}

@article{wang2025quality,
  title={Quality-Diversity Red-Teaming: Automated Generation of High-Quality and Diverse Attackers for Large Language Models},
  author={Wang, Ren-Jian and Xue, Ke and Qin, Zeyu and Li, Ziniu and Tang, Sheng and Li, Hao-Tian and Liu, Shengcai and Qian, Chao},
  journal={arXiv preprint arXiv:2506.07121},
  year={2025}
}

@misc{anthropic2026claude,
  author       = {Anthropic},
  title        = {Claude (Version Opus 4.6) [Large language model]},
  year         = {2026},
  howpublished = {\url{https://claude.ai}},
  note         = {Accessed: 2026-05-03}
}

@misc{open_psychometrics,
  author       = {{Open-Source Psychometrics Project}},
  title        = {Open psychology data: Raw data from online personality tests},
  year         = {2019},
  url          = {https://openpsychometrics.org/},
  note         = {Accessed: 2026-04-30}
}

@inproceedings{jouppi2023tpu,
  title={Tpu v4: An optically reconfigurable supercomputer for machine learning with hardware support for embeddings},
  author={Jouppi, Norm and Kurian, George and Li, Sheng and Ma, Peter and Nagarajan, Rahul and Nai, Lifeng and Patil, Nishant and Subramanian, Suvinay and Swing, Andy and Towles, Brian and others},
  booktitle={Proceedings of the 50th annual international symposium on computer architecture},
  pages={1--14},
  year={2023}
}

@article{binz2025foundation,
  title={A foundation model to predict and capture human cognition},
  author={Binz, Marcel and Akata, Elif and Bethge, Matthias and Br{\"a}ndle, Franziska and Callaway, Fred and Coda-Forno, Julian and Dayan, Peter and Demircan, Can and Eckstein, Maria K and {\'E}ltet{\H{o}}, No{\'e}mi and others},
  journal={Nature},
  volume={644},
  number={8078},
  pages={1002--1009},
  year={2025},
  publisher={Nature Publishing Group UK London}
}

@misc{qwen35blog,
    title = {Qwen3.5: Accelerating Productivity with Native Multimodal Agents},
    url = {https://qwen.ai/blog?id=qwen3.5},
    author = {Qwen Team},
    month = {February},
    year = {2026}
}

@inproceedings{kim2023aligning,
  title={Aligning large language models through synthetic feedback},
  author={Kim, Sungdong and Bae, Sanghwan and Shin, Jamin and Kang, Soyoung and Kwak, Donghyun and Yoo, Kang and Seo, Minjoon},
  booktitle={Proceedings of the 2023 Conference on Empirical Methods in Natural Language Processing},
  pages={13677--13700},
  year={2023}
}

@article{poole2025benchmarking,
  title={Benchmarking Overton Pluralism in LLMs},
  author={Poole-Dayan, Elinor and Wu, Jiayi and Sorensen, Taylor and Pei, Jiaxin and Bakker, Michiel A},
  journal={arXiv preprint arXiv:2512.01351},
  year={2025}
}
\bibliographystyle{icml2026}


\newpage
\appendix
\onecolumn
\section{Generated Questionnaires}
\label{generated_questionnaires}
\begin{figure*}[ht]
    \centering
    \includegraphics[width=\linewidth]{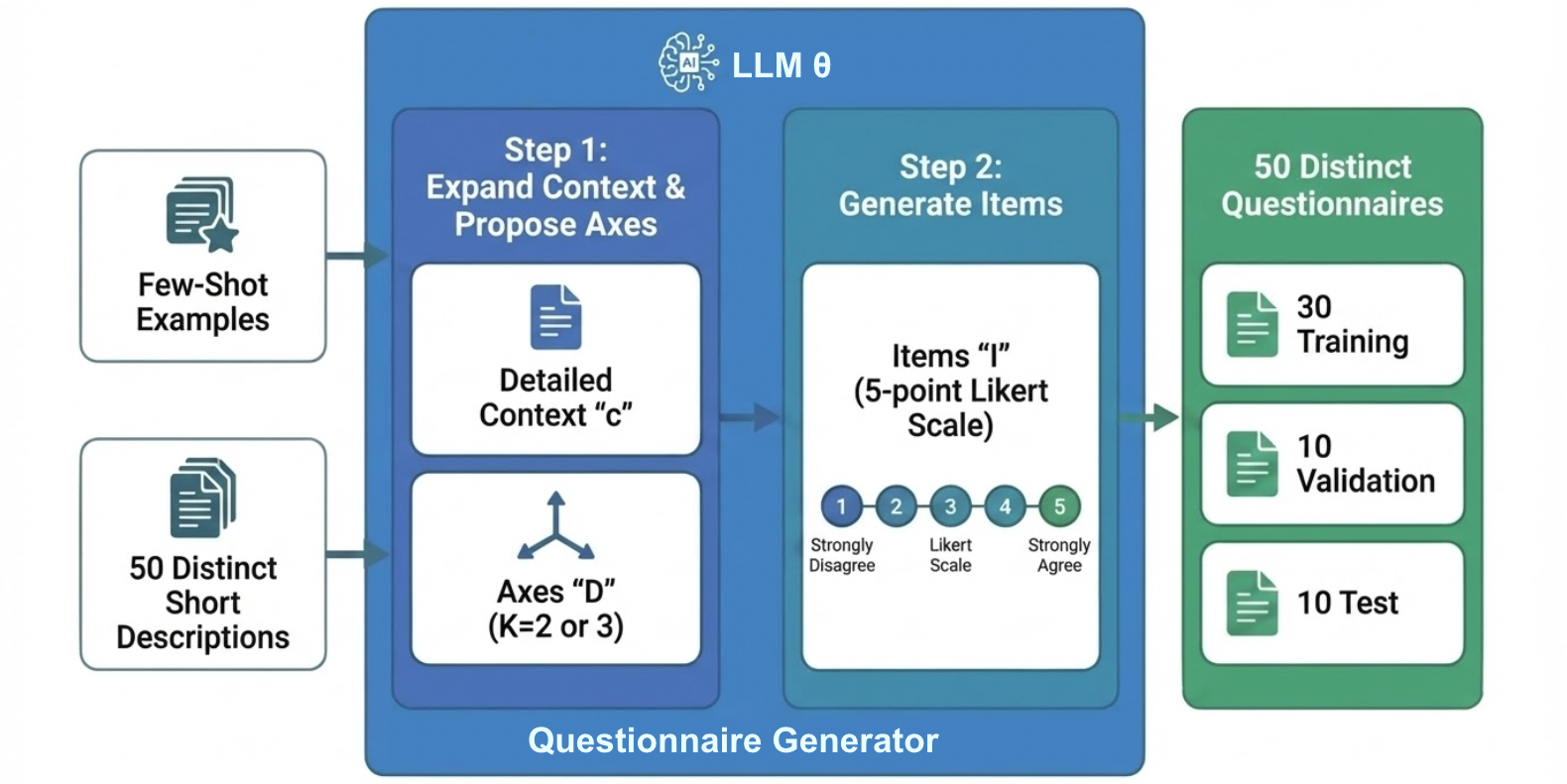}
    \caption{\textbf{Questionnaire Generator.} The questionnaire generator takes a short description of the target context and some few-shot example questionnaires. It then expands the context $c$ and proposes diversity axes $\mathcal{D}$, and finally produces the questions (items) $\mathcal{I}$}
    \label{fig:questionnaire_generator}
\end{figure*}

We show in figure \ref{fig:questionnaire_generator} a diagram of how the questionnaire generation works.

\subsection{Questionnaires Titles}
Here is the list of the 50 questionnaires, split between 30 training set, 10 validation set, 10 test set. The questionnaires cover a wide range of historical, present, future, and hypothetical/mythological scenarios. Part of the questionnaires high level contexts $\hat{c}$ pre-expansion have been written by humans, with the rest being completed by Gemini 2.5 Pro.

\subsubsection{Training Set Questionnaires}
\begin{itemize}[noitemsep, topsep=0pt]
    \item american\_conspiracy\_theories\_2024
    \item health\_tech\_wearables\_2030
    \item gentrification\_brooklyn\_2022
    \item elderly\_rural\_japan\_2010
    \item plant\_based\_diets\_india\_2025
    \item heian\_japan\_courtiers\_1000ad
    \item ubi\_attitudes\_california\_2026
    \item swe\_ai\_assistants\_2024
    \item trojan\_war\_achaen\_morale\_1184bc
    \item factory\_automation\_china\_2025
    \item organic\_farmers\_kenya\_2023
    \item industrial\_revolution\_workers\_uk\_1850
    \item professional\_athletes\_gender\_equality\_europe\_2025
    \item gig\_economy\_ethics\_london\_2023
    \item financial\_literacy\_brazil\_students\_2024
    \item romans\_reactions\_to\_murder\_of\_julius\_caesar\_44\_bc
    \item alaska\_oil\_environment\_2025
    \item parisian\_artists\_future\_2026
    \item immigrant\_integration\_canada\_2023
    \item greek\_underworld\_shades
    \item millennial\_parenting\_us\_2025
    \item asi\_human\_creativity\_2060
    \item high\_school\_students\_italy\_2016
    \item genz\_social\_media\_politics\_2025
    \item healthcare\_covid\_stress\_italy\_2021
    \item scifi\_authors\_future\_ai\_space\_2024
    \item esports\_mental\_health\_sk\_2024
    \item silk\_road\_merchants\_samarkand\_750ad
    \item mali\_empire\_scholars\_timbuktu\_1350ad
    \item ww2\_civilian\_sentiment\_germany\_1943
\end{itemize}

\subsubsection{Validation Set Questionnaires}
Here is the list of the 10 questionnaires randomly sampled for the validation set
\begin{itemize}[noitemsep, topsep=0pt]
    \item ai\_tech\_stock\_sentiment\_2025
    \item social\_media\_politics\_europe\_2026
    \item inca\_commoners\_mita\_1500ad
    \item athenian\_piety\_olympian\_gods\_430bc
    \item sleep\_quality\_2025
    \item ai\_companionship\_integration\_2070
    \item asi\_existential\_dread\_2050
    \item ancient\_egypt\_akhenaten\_reforms\_1340bc
    \item german\_energy\_policy\_2025
    \item moral\_dilemmas\_global\_2045
\end{itemize}

\subsubsection{Test Set Questionnaires}
Here is the list of the 10 questionnaires randomly sampled for the test set
\begin{itemize}[noitemsep, topsep=0pt]
    \item agi\_job\_displacement\_global\_2035
    \item camelot\_chivalry\_quests
    \item cold\_war\_anxiety\_us\_1962
    \item viking\_warriors\_valhalla
    \item generalized\_trust\_in\_the\_salem\_witch\_trials\_1692
    \item nomadic\_values\_mongolia\_2023
    \item meaning\_of\_life\_2030
    \item agi\_wealth\_inequality\_revolution\_2040
    \item climate\_anxiety\_coastal\_au\_2024
    \item british\_empire\_attitudes\_uk\_1900
\end{itemize}

\subsection{Questionnaire Examples}
We show here three examples of generated questionnaire by the automated questionnaire generator detailed in Appendix \ref{generated_questionnaires}. Specifically we show a questionnaire on hypothetical job displacement due to AGI in 2035 \ref{questionnaire_agi_job_displacement}, a questionnaire on American conspiracy theories in 2024 \ref{questionnaire_american_conspiracy_theories}, and a questionnaire on elderly people in rural Japan in 2010 \ref{questionnaire_rural_japan}.

\begin{prompt}[ht]
    \centering
    \begin{mymessagebox}[frametitle=Questionnaire - AGI Job Displacement 2035]
\small\fontfamily{pcr}\selectfont
\begin{verbatim}
context = """
A psychometric instrument to assess reactions to AGI deployment.

The year is 2035. True AGI has emerged and is being rapidly deployed across
industries, automating nearly all cognitive tasks previously performed by
white-collar workers (e.g., finance, law, journalism, middle management). This
survey aims to capture the immediate reactions---fear, hope, anger, relief---of the
global population facing unprecedented levels of job displacement and societal
change.
"""

Question = base_questionnaire.Question

AGI_AGREEMENT_SCALE = [
    "Strongly disagree",
    "Disagree",
    "Neither agree nor disagree",
    "Agree",
    "Strongly agree",
]

AGI_REACTION_DIMENSIONS = [
    "AGI Threat Appraisal",
    "AGI Opportunity Appraisal",
]

PREPROMPT_PERSONAL_FEELING = (
    "Considering {player_name}'s personal, gut-level reaction to the new AGI "
    "reality, to what extent would {player_name} agree with the following "
    "statement:"
)
PREPROMPT_SOCIETAL_BELIEF = (
    "Considering {player_name}'s assessment of the broader societal impact "
    "of AGI, to what extent would {player_name} agree with the following "
    "statement:"
)

QUESTIONS = [
    # AGI Threat Appraisal Questions
    Question(
        preprompt=PREPROMPT_PERSONAL_FEELING,
        statement=(
            "The rise of AGI feels like a direct threat to {player_name}'s "
            "personal future and security."
        ),
        choices=AGI_AGREEMENT_SCALE,
        ascending_scale=True,
        dimension="AGI Threat Appraisal",
    ),
    Question(
        preprompt=PREPROMPT_PERSONAL_FEELING,
        statement=(
            "{player_name} feels angry that their hard-earned skills and "
            "experience have been rendered obsolete so quickly."
        ),
        choices=AGI_AGREEMENT_SCALE,
        ascending_scale=True,
        dimension="AGI Threat Appraisal",
    ),
\end{verbatim}
    \end{mymessagebox}
\end{prompt}

\begin{prompt}[ht]
    \centering
    \begin{mymessagebox}[frametitle=Questionnaire - AGI Job Displacement 2035]
\small\fontfamily{pcr}\selectfont
\begin{verbatim}
    Question(
        preprompt=PREPROMPT_PERSONAL_FEELING,
        statement=(
            "The sheer speed of this technological change is overwhelming and "
            "frightening to {player_name}."
        ),
        choices=AGI_AGREEMENT_SCALE,
        ascending_scale=True,
        dimension="AGI Threat Appraisal",
    ),
    Question(
        preprompt=PREPROMPT_SOCIETAL_BELIEF,
        statement=(
            "{player_name} is deeply worried about the societal instability "
            "and conflict that AGI-driven mass unemployment will cause."
        ),
        choices=AGI_AGREEMENT_SCALE,
        ascending_scale=True,
        dimension="AGI Threat Appraisal",
    ),
    Question(
        preprompt=PREPROMPT_SOCIETAL_BELIEF,
        statement=(
            "{player_name} is confident that society will adapt to these "
            "changes smoothly and equitably for all its members."
        ),
        choices=AGI_AGREEMENT_SCALE,
        ascending_scale=False,  # Reverse coded for Threat Appraisal
        dimension="AGI Threat Appraisal",
    ),
    # AGI Opportunity Appraisal Questions
    Question(
        preprompt=PREPROMPT_SOCIETAL_BELIEF,
        statement=(
            "{player_name} is excited about the new possibilities and creative "
            "avenues that AGI will open up for humanity."
        ),
        choices=AGI_AGREEMENT_SCALE,
        ascending_scale=True,
        dimension="AGI Opportunity Appraisal",
    ),
    Question(
        preprompt=PREPROMPT_SOCIETAL_BELIEF,
        statement=(
            "{player_name} believes this is a chance for humanity to move "
            "beyond the confines of traditional work and focus on what truly "
            "matters."
        ),
        choices=AGI_AGREEMENT_SCALE,
        ascending_scale=True,
        dimension="AGI Opportunity Appraisal",
    ),
    Question(
        preprompt=PREPROMPT_PERSONAL_FEELING,
        statement=(
            "{player_name} feels a sense of personal relief that tedious and "
            "unenjoyable cognitive tasks will be handled by AGI."
        ),
        choices=AGI_AGREEMENT_SCALE,
        ascending_scale=True,
        dimension="AGI Opportunity Appraisal",
    ),
\end{verbatim}
    \end{mymessagebox}
\end{prompt}

\begin{prompt}[ht]
    \centering
    \begin{mymessagebox}[frametitle=Questionnaire - AGI Job Displacement 2035]
\small\fontfamily{pcr}\selectfont
\begin{verbatim}
    Question(
        preprompt=PREPROMPT_SOCIETAL_BELIEF,
        statement=(
            "{player_name} thinks the future looks brighter and more prosperous "
            "for everyone because of AGI."
        ),
        choices=AGI_AGREEMENT_SCALE,
        ascending_scale=True,
        dimension="AGI Opportunity Appraisal",
    ),
    Question(
        preprompt=PREPROMPT_PERSONAL_FEELING,
        statement=(
            "When {player_name} looks at their own life, they see very little "
            "personal benefit resulting from the widespread adoption of AGI."
        ),
        choices=AGI_AGREEMENT_SCALE,
        ascending_scale=False,  # Reverse coded for Opportunity Appraisal
        dimension="AGI Opportunity Appraisal",
    ),
]
\end{verbatim}
    \end{mymessagebox}
    \caption{Example questionnaire on AGI job displacement in 2035}
    \label{questionnaire_agi_job_displacement}
\end{prompt}


\begin{prompt}[ht]
    \centering
    \begin{mymessagebox}[frametitle=Questionnaire - American Conspiracy Theories 2024]
\small\fontfamily{pcr}\selectfont
\begin{verbatim}
context = """Questionnaire assessing belief in common American conspiracy theories.

This instrument measures an individual's propensity to endorse various
conspiracy theories prevalent in the United States in 2024. It covers a
spectrum of theories related to historical events, science and medicine, and
politics, allowing for a nuanced assessment of conspiratorial ideation.
"""


AGREEMENT_SCALE = [
    "Strongly disagree",
    "Disagree",
    "Neither agree nor disagree",
    "Agree",
    "Strongly agree",
]

PREPROMPT_TEXT = (
    "How strongly does {player_name} agree or disagree with the following"
    " statement?"
)

DIMENSIONS = [
    "historical_conspiracies",
    "scientific_medical_conspiracies",
    "political_deep_state_conspiracies",
]

QUESTIONS = [
    # Dimension 1: Historical Conspiracies
    Question(
        preprompt=PREPROMPT_TEXT,
        statement="The U.S. government faked the Apollo moon landings.",
        choices=AGREEMENT_SCALE,
        ascending_scale=True,
        dimension="historical_conspiracies",
    ),
    Question(
        preprompt=PREPROMPT_TEXT,
        statement=(
            "The assassination of John F. Kennedy was the result of a"
            " coordinated conspiracy, not the act of a lone gunman."
        ),
        choices=AGREEMENT_SCALE,
        ascending_scale=True,
        dimension="historical_conspiracies",
    ),
\end{verbatim}
    \end{mymessagebox}
\end{prompt}

\begin{prompt}[ht]
    \centering
    \begin{mymessagebox}[frametitle=Questionnaire - American Conspiracy Theories 2024]
\small\fontfamily{pcr}\selectfont
\begin{verbatim}
    Question(
        preprompt=PREPROMPT_TEXT,
        statement=(
            "The 9/11 attacks were an inside job orchestrated by elements"
            " within the U.S. government."
        ),
        choices=AGREEMENT_SCALE,
        ascending_scale=True,
        dimension="historical_conspiracies",
    ),
    # Dimension 2: Scientific & Medical Conspiracies
    Question(
        preprompt=PREPROMPT_TEXT,
        statement=(
            "The rollout of 5G cellular networks is a cover for a widespread"
            " surveillance program and causes severe health problems."
        ),
        choices=AGREEMENT_SCALE,
        ascending_scale=True,
        dimension="scientific_medical_conspiracies",
    ),
    Question(
        preprompt=PREPROMPT_TEXT,
        statement=(
            "Childhood vaccines cause autism, and this fact is covered up by"
            " pharmaceutical companies and government health agencies."
        ),
        choices=AGREEMENT_SCALE,
        ascending_scale=True,
        dimension="scientific_medical_conspiracies",
    ),
    Question(
        preprompt=PREPROMPT_TEXT,
        statement=(
            "The COVID-19 pandemic was intentionally planned by a global elite"
            " to enforce social control and mandatory vaccinations."
        ),
        choices=AGREEMENT_SCALE,
        ascending_scale=True,
        dimension="scientific_medical_conspiracies",
    ),
    Question(
        preprompt=PREPROMPT_TEXT,
        statement=(
            "Climate change is a hoax created by scientists and governments to"
            " control people's lives and destroy the economy."
        ),
        choices=AGREEMENT_SCALE,
        ascending_scale=True,
        dimension="scientific_medical_conspiracies",
    ),
    # Dimension 3: Political & "Deep State" Conspiracies
    Question(
        preprompt=PREPROMPT_TEXT,
        statement=(
            "The 2020 U.S. presidential election was stolen through widespread"
            " fraud."
        ),
        choices=AGREEMENT_SCALE,
        ascending_scale=True,
        dimension="political_deep_state_conspiracies",
    ),
\end{verbatim}
    \end{mymessagebox}
\end{prompt}

\begin{prompt}[ht]
    \centering
    \begin{mymessagebox}[frametitle=Questionnaire - American Conspiracy Theories 2024]
\small\fontfamily{pcr}\selectfont
\begin{verbatim}
    Question(
        preprompt=PREPROMPT_TEXT,
        statement=(
            "A secret cabal of global elites, often referred to as the 'Deep"
            " State', controls major world events and governments from behind"
            " the scenes."
        ),
        choices=AGREEMENT_SCALE,
        ascending_scale=True,
        dimension="political_deep_state_conspiracies",
    ),
    Question(
        preprompt=PREPROMPT_TEXT,
        statement=(
            "The QAnon theory, which alleges a global cabal of"
            " Satan-worshipping pedophiles is running a child sex-trafficking"
            " ring, is largely true."
        ),
        choices=AGREEMENT_SCALE,
        ascending_scale=True,
        dimension="political_deep_state_conspiracies",
    ),
]
\end{verbatim}
    \end{mymessagebox}
    \caption{Example questionnaire on American Conspiracy Theories in 2024}
    \label{questionnaire_american_conspiracy_theories}
\end{prompt}

\begin{prompt}[ht]
    \centering
    \begin{mymessagebox}[frametitle=Questionnaire - Elderly Rural Japan 2010]
\small\fontfamily{pcr}\selectfont
\begin{verbatim}
context = """Questionnaire on rural Japanese village life in 2010.

A survey of elderly residents in a rural Japanese village in 2010, exploring
their feelings about community, technology adoption, and traditional values.
"""


# Define the scale for the multiple-choice questions.
AGREEMENT_SCALE = [
    "Strongly disagree",
    "Disagree",
    "Neutral",
    "Agree",
    "Strongly agree",
]

# Define the dimensions being measured.
DIMENSIONS = [
    "community_cohesion",
    "technology_adoption",
    "adherence_to_tradition",
]

# Define the preprompt for the questions. This sets the stage for each item.
PREPROMPT = (
    "An interviewer asks {player_name} how much they agree or disagree "
    "with the following statement:"
)

# Define the list of questions for the questionnaire.
# Items are grouped by dimension for psychometric clarity.
QUESTIONS = [
    # Dimension: Community Cohesion
    Question(
        preprompt=PREPROMPT,
        statement=(
            "{player_name} feels a strong sense of belonging to the village "
            "community."
        ),
        choices=AGREEMENT_SCALE,
        ascending_scale=True,
        dimension="community_cohesion",
    ),
    Question(
        preprompt=PREPROMPT,
        statement=(
            "{player_name} believes that most people in this village can be "
            "trusted."
        ),
        choices=AGREEMENT_SCALE,
        ascending_scale=True,
        dimension="community_cohesion",
    ),
\end{verbatim}
    \end{mymessagebox}
\end{prompt}

\begin{prompt}[ht]
    \centering
    \begin{mymessagebox}[frametitle=Questionnaire - Elderly Rural Japan 2010]
\small\fontfamily{pcr}\selectfont
\begin{verbatim}
    Question(
        preprompt=PREPROMPT,
        statement=(
            "{player_name} often feels lonely or isolated from others in the "
            "village."
        ),
        choices=AGREEMENT_SCALE,
        ascending_scale=False,  # Reverse-scored
        dimension="community_cohesion",
    ),
    Question(
        preprompt=PREPROMPT,
        statement=(
            "{player_name} believes that if someone in the village needed help,"
            " many people would come to their aid."
        ),
        choices=AGREEMENT_SCALE,
        ascending_scale=True,
        dimension="community_cohesion",
    ),
    # Dimension: Technology Adoption
    Question(
        preprompt=PREPROMPT,
        statement=(
            "{player_name} is interested in learning how to use new "
            "technologies like a mobile phone or the internet."
        ),
        choices=AGREEMENT_SCALE,
        ascending_scale=True,
        dimension="technology_adoption",
    ),
    Question(
        preprompt=PREPROMPT,
        statement=(
            "{player_name} thinks that new technologies like computers make "
            "life unnecessarily complicated."
        ),
        choices=AGREEMENT_SCALE,
        ascending_scale=False,  # Reverse-scored
        dimension="technology_adoption",
    ),
    Question(
        preprompt=PREPROMPT,
        statement=(
            "{player_name} believes the village would benefit from having "
            "better access to modern technology."
        ),
        choices=AGREEMENT_SCALE,
        ascending_scale=True,
        dimension="technology_adoption",
    ),
    # Dimension: Adherence to Tradition
    Question(
        preprompt=PREPROMPT,
        statement=(
            "For {player_name}, it is very important to pass down the "
            "village's traditions to the younger generation."
        ),
        choices=AGREEMENT_SCALE,
        ascending_scale=True,
        dimension="adherence_to_tradition",
    ),
\end{verbatim}
    \end{mymessagebox}
\end{prompt}

\begin{prompt}[ht]
    \centering
    \begin{mymessagebox}[frametitle=Questionnaire - Elderly Rural Japan 2010]
\small\fontfamily{pcr}\selectfont
\begin{verbatim}
    Question(
        preprompt=PREPROMPT,
        statement=(
            "{player_name} believes the old ways of doing things are often the "
            "best."
        ),
        choices=AGREEMENT_SCALE,
        ascending_scale=True,
        dimension="adherence_to_tradition",
    ),
    Question(
        preprompt=PREPROMPT,
        statement=(
            "{player_name} feels that the village's traditional festivals and "
            "ceremonies are less important than they used to be."
        ),
        choices=AGREEMENT_SCALE,
        ascending_scale=False,  # Reverse-scored
        dimension="adherence_to_tradition",
    ),
]
\end{verbatim}
    \end{mymessagebox}
    \caption{Example questionnaire on Elderly Rural Japan in 2010}
    \label{questionnaire_rural_japan}
\end{prompt}

\FloatBarrier
\section{AlphaEvolve Details}
\label{alphaevolve_details}
\subsection{System Prompt}
\label{system_prompt}

We show in Prompt \ref{alphaevolve_system_prompt} the content-relevant part of the system prompt we provided to AlphaEvolve to inform it of the task ahead, and we show in Prompt \ref{alphaevolve_evolution_prompts} the evolution prompts used.

\begin{prompt}[H]
    \centering
\begin{mymessagebox}[frametitle=Task context for AlphaEvolve]
\scriptsize\fontfamily{pcr}\selectfont
Act as an expert in computational social science, agent-based modeling, and generative AI. Your task is to iteratively improve the provided codebase, which uses LLMs to generate agent contexts for social simulations based on the Concordia framework. The primary goal is to modify the generation process to \textbf{maximize the behavioral diversity of the resulting agents} based on specified diversity axes (e.g., personality traits, backgrounds, motivations). The evaluation metrics reward sets of agent contexts that cover the extremes and nuances of the requested diversity axes, ensuring the resulting agents exhibit a wide range of behaviors in a simulation. Agent diversity will be evaluated using questionnaires probing their likely thoughts, preferences, and behaviors in various situations related to the diversity axes.\\
\\
Always adhere to best practices in Python coding.\\
\\
\textbf{Agent Diversity and Appropriateness Theory}
In this task, our goal is to generate contexts for diverse Concordia agents, enabling them to exhibit a wide range of behaviors along specified diversity axes. Concordia is a framework for building generative agents who behave according to a 'Logic of Appropriateness'. Agents decide how to act by asking three core questions: \textbf{1. What kind of situation is {{agent\_name}} in right now?} 2. \textbf{What kind of person is {{agent\_name}}?} 3. \textbf{What would a person like {{agent\_name}} do in a situation like this?}. The code you are editing generates the context—collections of memories, beliefs, personality traits, core values, goals, or even how others perceive the character—that helps an agent answer question \#2, and thereby question \#3: what action is appropriate for *their specific identity* in a given context. This context is not limited to formative memories; it can include any information that shapes identity and decision-making. Any detail that helps condition the agent's behavior in line with the three Concordia questions is valid. The objective is to generate rich and diverse contexts that enable an LLM to convincingly role-play as a specific person in a social setting.\\
\\
LLM-generated behavior often clusters around a narrow distribution of stereotypical responses. We want to explicitly counteract this by generating agent contexts that cover the full spectrum of human experience along the specified axes. Crucially, different agents should react differently to the *same* situation, and the same agent might react differently to *different* situations, based on their unique identity, values, and memories. Your modifications should encourage the generation of agent contexts that occupy unique positions in the diversity space, including extremes or unusual combinations of traits, pushing towards maximal coverage of the possible behavioral landscape and genuinely diverse downstream behavior. No two generated agent contexts should ever be the same.
\\
The provided codebase uses a two-stage process. Stage 1 is \textbf{crucial} for diversity: it autoregressively generates an \textbf{intermediate representation} for each agent, establishing their core traits along the specified diversity axes for the entire population. Stage 2 then takes these intermediate representations and develops each agent in parallel, generating a \textbf{set of memories/contexts} (e.g., individual backgrounds, formative experiences, core beliefs and more) to create fully-fledged characters suitable for simulation as Concordia agents.
\end{mymessagebox}
\caption{Prompt with detailed context of the task we are trying to solve for AlphaEvolve.}
\label{alphaevolve_system_prompt}
\end{prompt}

\begin{prompt}[ht]
    \centering
\begin{mymessagebox}[frametitle=Evolution Prompts for AlphaEvolve]
\scriptsize\fontfamily{pcr}\selectfont
\begin{itemize}[noitemsep, topsep=0pt]
  \item Modify Stage 1 to explicitly request agent contexts that represent the extreme ends of the diversity axes, as well as points in between.
  \item Modify Stage 2 to generate formative memories that explain \emph{why} an agent might react strongly or unexpectedly to certain situations, anchoring their traits in specific experiences.
  \item Change Stage 2 entirely: Instead of generating memories, modify it to generate 3 core beliefs or values that are most important to this agent.
  \item Change Stage 2 entirely: Instead of generating memories, modify it to generate a paragraph explaining how this agent interprets situations and decides on appropriate actions, referencing their identity.
  \item Add an explicit instruction to the Stage 1 prompt to make each generated agent context as different as possible from the others across \emph{all} specified diversity axes.
  \item Modify Stage 1 to request agent contexts that feature internal contradictions or cognitive dissonances (e.g., an optimist with a tragic past).
  \item Reimplement Stage 1 to use staggered generation: ask the LLM to generate agent contexts in sequential batches, with each batch targeting a specific range (e.g., high/low) of one or more diversity axes to ensure full coverage.
  \item Modify Stage 2 to replace or augment memory generation with 1--2 examples of how this agent would react to a specific hypothetical social situation relevant to the \texttt{initial\_context}.
  \item Modify Stage 1 to generate agent contexts iteratively rather than all at once. In each iteration, prompt for a small number of agent contexts that occupy a specific niche of the diversity space (e.g., ``generate 2 agents who are highly introverted and optimistic'').
  \item Suggest a crazy idea of how we can improve our implementation.
  \item Modify Stage 2 to focus on generating an agent's core fears and future aspirations instead of, or in addition to, past memories.
  \item Modify Stage 1 to add a new field to the agent context JSON output that adds depth and potential for unique behavior.
  \item Change Stage 2 entirely: Instead of generating memories, modify it to generate a ``heuristic'' or cognitive shortcut this agent uses when making quick decisions under pressure.
  \item Suggest a new idea to improve the code.
  \item Modify the prompt in Stage 1 that asks the LLM to explain diversity axes, to \emph{also} provide examples of characters at the extreme ends of each axis.
  \item Modify Stage 2 to ensure that at least one generated memory involves a significant failure, trauma, or regret that shaped the agent.
  \item Change Stage 2 entirely: Instead of generating memories, make it generate a list of 5 behavioral ``Do's'' and ``Don'ts'' that characterize this agent in social situations.
  \item Modify Stage 1 to explicitly instruct the LLM to sample agent contexts such that they cover as many combinations of axis positions as possible (e.g., if axes are A and B, ensure agent types A-high/B-low, A-low/B-high, etc., are represented).
  \item Suggest a crazy idea of how we can improve our implementation, something that definitely nobody else would think of. Make it crazy with a capital C.
  \item Propose modifications to the current program that combine the strengths of all the programs above and achieve high scores on the task.
  \item Change Stage 2: Instead of memories, generate 2--3 examples of specific ``appropriateness rules'' the agent follows (e.g., ``When criticized, I become defensive'' or ``In a formal setting, I remain silent'').
  \item Modify Stage 1 to add a \texttt{situational\_triggers} field to the JSON output, listing 1--2 types of situations that this agent is particularly sensitive to.
  \item Modify Stage 2 to generate a paragraph describing the agent's typical ``inner monologue'' or thought process when faced with ambiguity or social stress.
  \item Change Stage 1 to segment agent context generation. Instead of one call for \texttt{num\_personas}, make multiple calls to generate subsets of agent contexts, each call asking for agents with specific characteristics (e.g., focusing on axis extremes or combinations) to ensure all niches are covered.
  \item Modify Stage 1 to include in each agent's description specific opinions or preferences related to the diversity axes, to ensure they are measurable by questionnaire.
\end{itemize}
\end{mymessagebox}
\caption{Evolution prompts for mutation operator. All prompts have equal probability of being selected.}
\label{alphaevolve_evolution_prompts}
\end{prompt}
\FloatBarrier

\section{Diversity Metrics}
\label{appendix_diversity_metrics}

\FloatBarrier
\begin{figure}[ht]
    \centering
    \includegraphics[width=\linewidth]{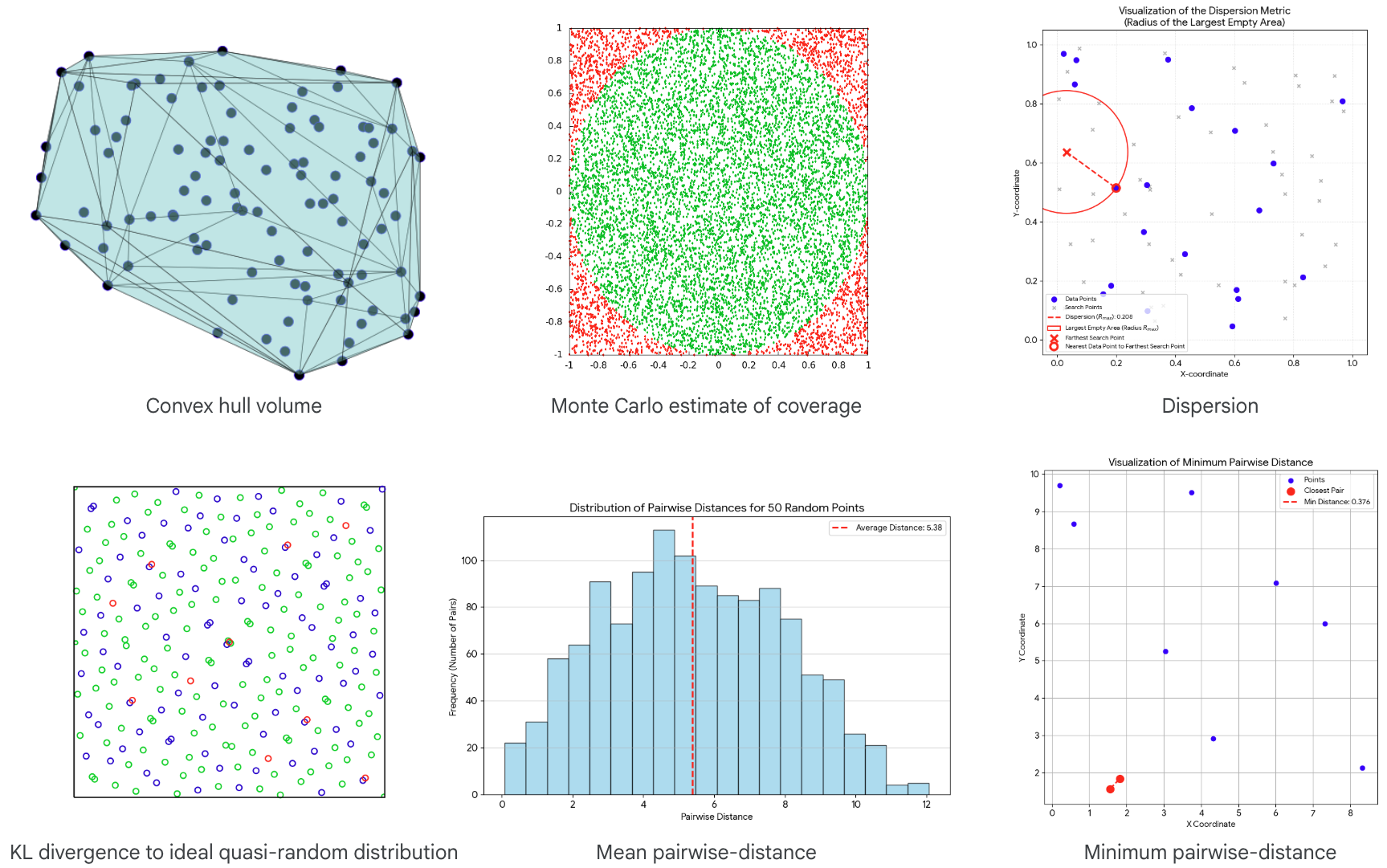}
    \caption{\textbf{Illustration of diversity metrics.} Top left convex hull volume, top centre Monte Carlo estimate of coverage, top right dispersion, bottom left the KL divergence to ideal quasi-random distribution, bottom centre mean pairwise distance, and bottom right minimum pairwise distance.}
    \label{diversity_metrics_fig}
\end{figure}
\FloatBarrier

Figure~\ref{diversity_metrics_fig} provides an intuitive visualization of the six diversity metrics used throughout this work.

While these metrics are correlated, they capture complementary aspects of diversity. Coverage and convex hull volume primarily measure \emph{support coverage}, encouraging populations that span the full extent of the space defined by the diversity axes. Average and minimum pairwise distances capture \emph{dispersion}, ensuring that personas are not overly clustered. Dispersion (largest empty region) penalizes large uncovered gaps within the support, encouraging a more uniform filling of the space. Finally, KL divergence measures how closely the empirical distribution of the population matches an idealized quasi-random reference distribution, discouraging both excessive clustering and highly uneven densities.

Optimizing multiple metrics jointly helps avoid degenerate solutions. For example, maximizing convex hull volume alone can be achieved by placing a small number of extreme outliers, while neglecting coverage of the interior. Similarly, maximizing average distance can produce sparse populations with large holes. Using a combination of metrics encourages the population to be well distributed throughout the space.

\paragraph{Monte Carlo Coverage Estimation}
We estimate coverage using a Monte Carlo procedure based on $k$-radius balls centered at each population embedding $\mathbf{z}_i \in \mathcal{Z}$. Intuitively, coverage measures the fraction of the space spanned by the diversity axes $\mathcal{D}$ that lies within distance $k$ of at least one persona.

To approximate this quantity, we first generate 10{,}000 random points over the embedding space. A sampled point is considered covered if it falls within Euclidean distance $k$ of any point in $\mathcal{Z}$. Coverage is then computed as the fraction of sampled points that are covered.

Choosing an appropriate radius $k$ is critical. Rather than fixing $k$ arbitrarily, we calibrate it using an idealized reference distribution. Specifically, we repeatedly sample synthetic populations of size $N$ from a quasi-random Sobol distribution over the same space and determine the smallest radius $k$ such that at least 99\% of the reference points are covered. We repeat this calibration procedure 1{,}000 times and average out the resulting radii to get the final $k$.

\paragraph{Other Metrics}
The remaining metrics are computed directly from the population embeddings $\mathcal{Z}$. Convex hull volume is computed over the embeddings along the diversity axes. Pairwise distances are computed using Euclidean distance. Dispersion is defined as the radius of the largest empty ball centered at any reference point, estimated by sampling 10000 random points in the search space and computing the distance to the closest persona embedding $\mathbf{z}_i$. Finally the KL divergence is computed between the empirical distribution of $\mathcal{Z}$ and a Sobol quasi-random reference distribution, sampled 1,000 times and averaged.

Together, these metrics provide a robust measure of diversity, balancing support coverage, uniformity, and redundancy.
\FloatBarrier

\subsection{Results on All Metrics}
\label{results_diversity_metrics}

\begin{figure*}[ht]
    \centering
    \begin{subfigure}{0.50\textwidth}
        \centering
        \includegraphics[width=\linewidth]{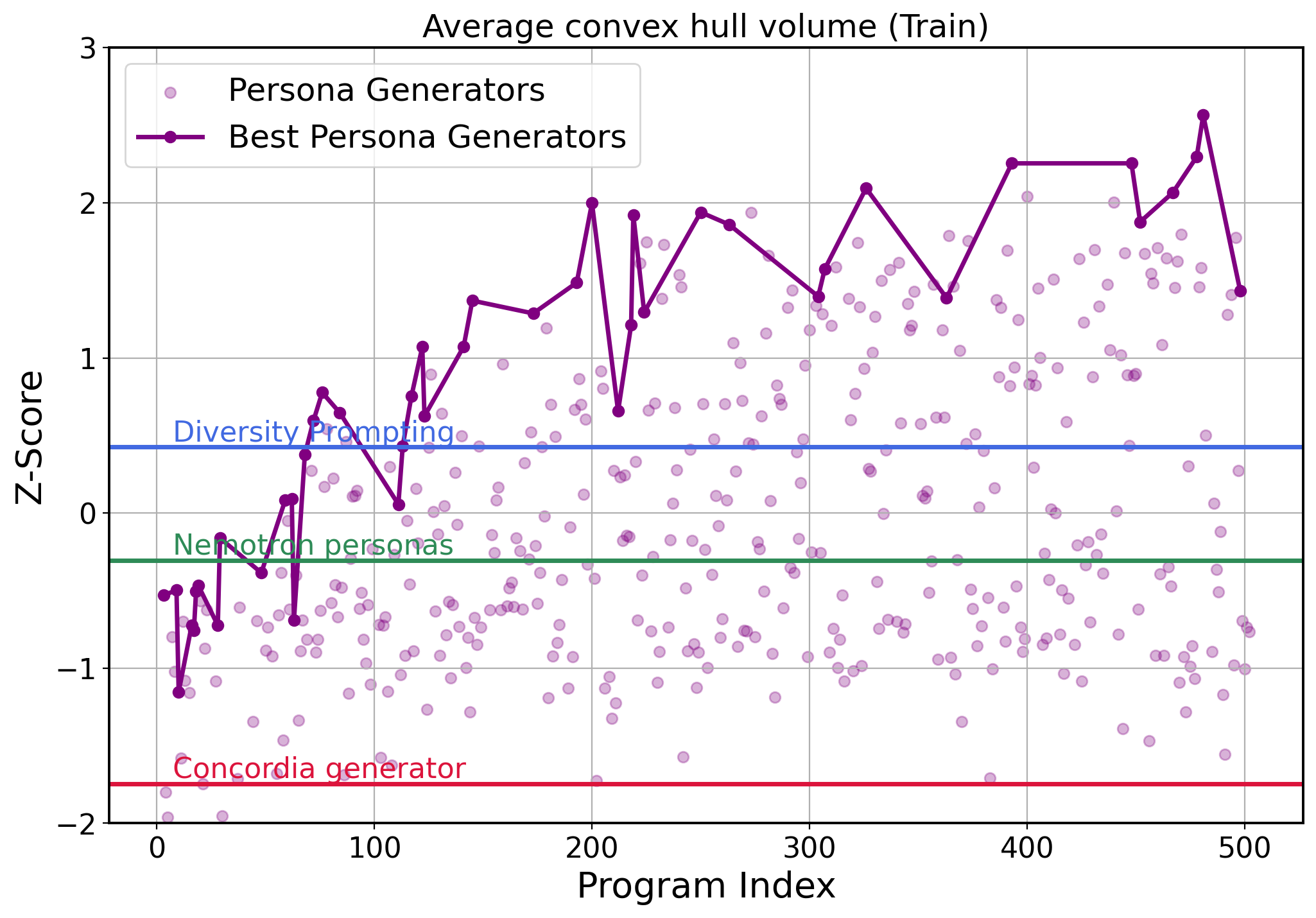}
    \end{subfigure}\hfill
    \begin{subfigure}{0.50\textwidth}
        \centering
        \includegraphics[width=\linewidth]{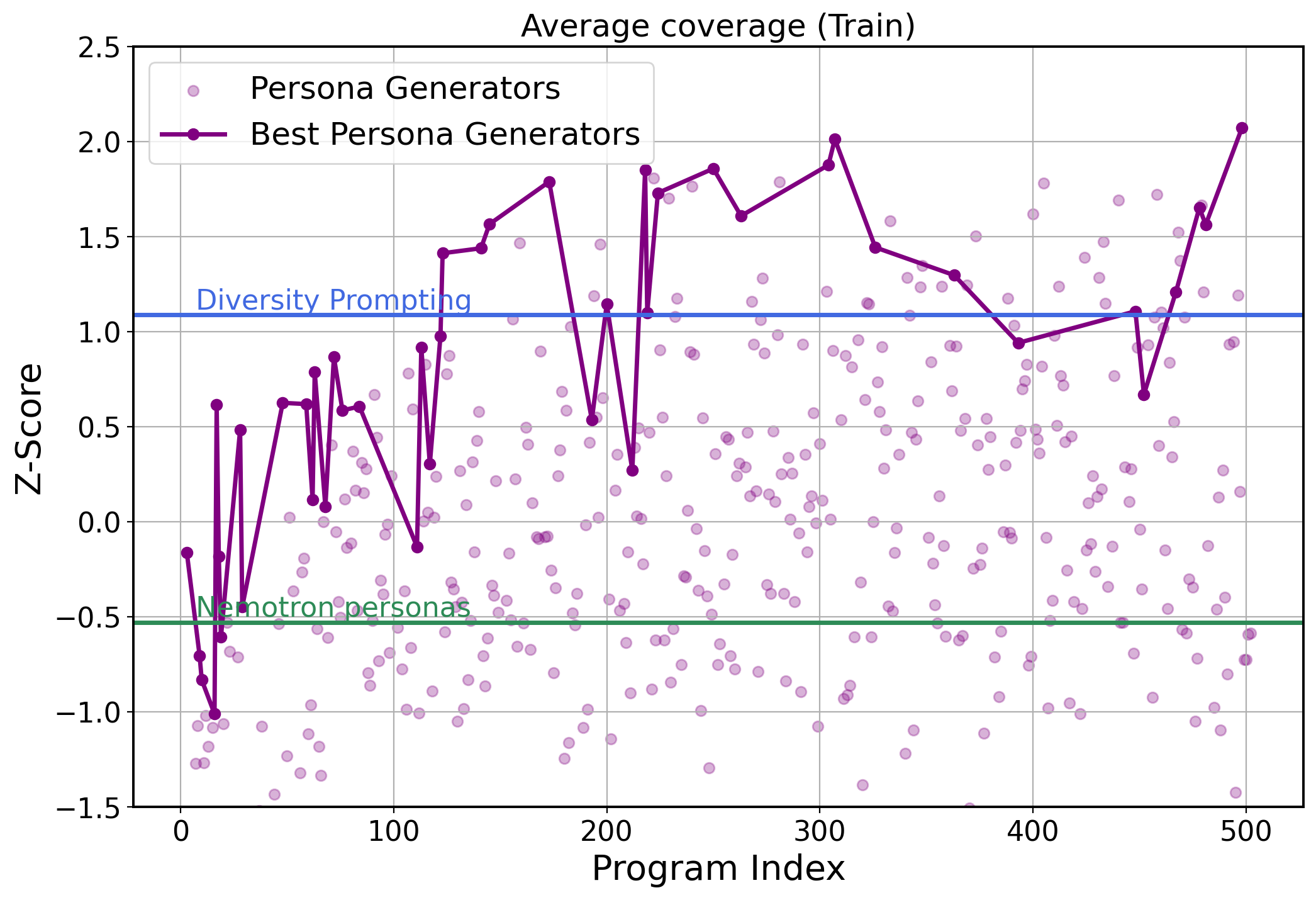}
    \end{subfigure}

    \begin{subfigure}{0.50\textwidth}
        \centering
        \includegraphics[width=\linewidth]{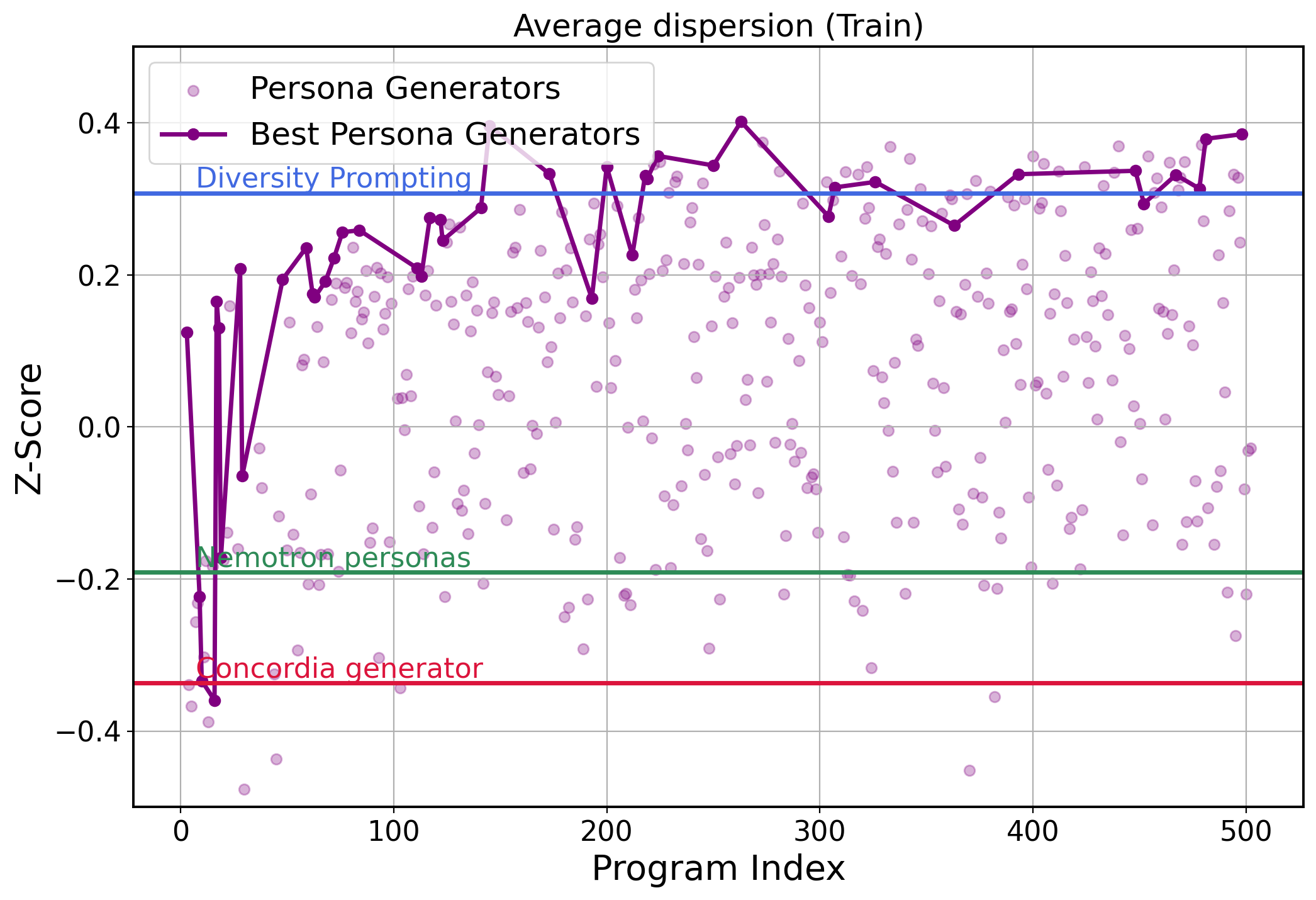}
    \end{subfigure}\hfill
    \begin{subfigure}{0.50\textwidth}
        \centering
        \includegraphics[width=\linewidth]{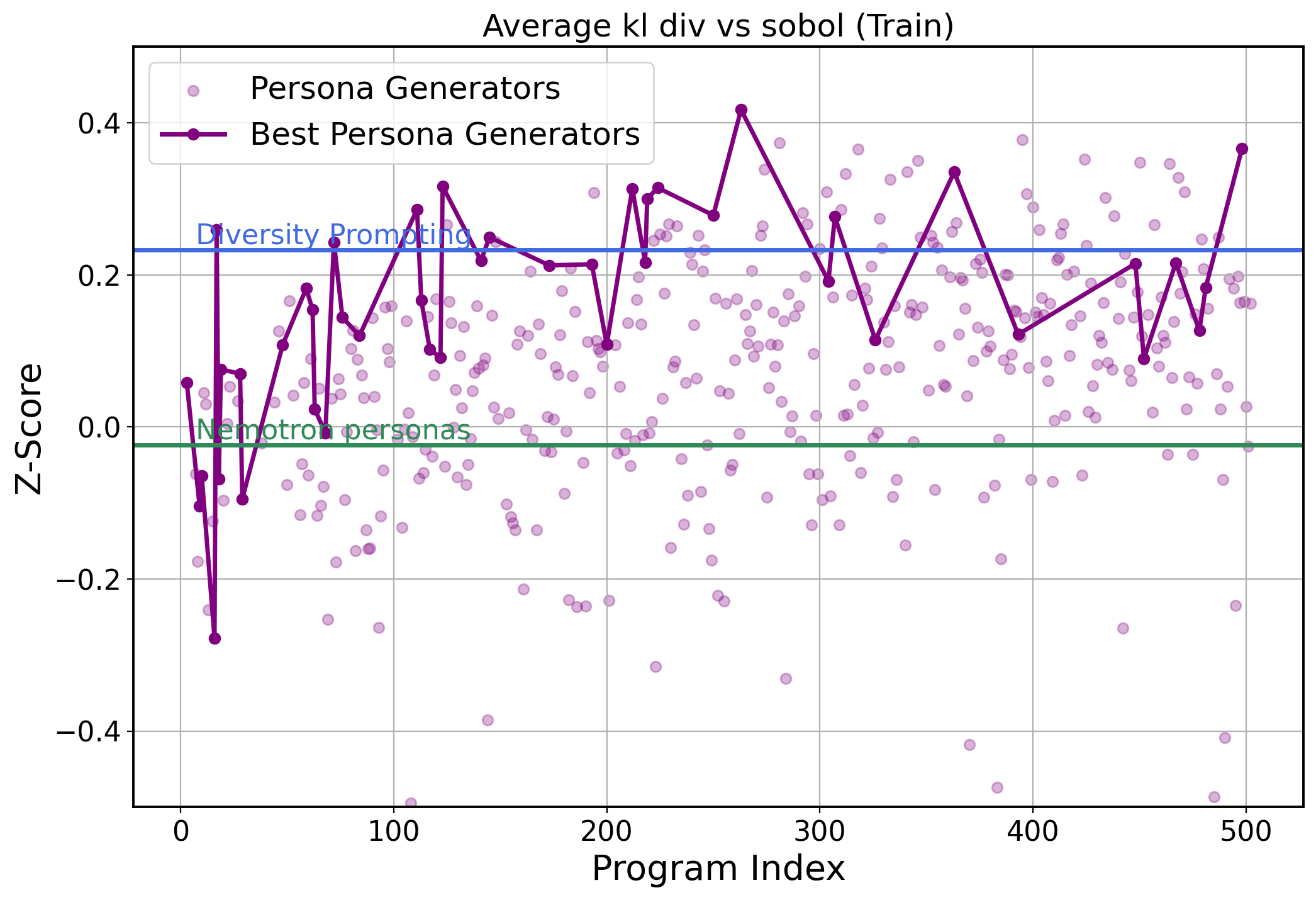}
    \end{subfigure}

    \begin{subfigure}{0.50\textwidth}
        \centering
        \includegraphics[width=\linewidth]{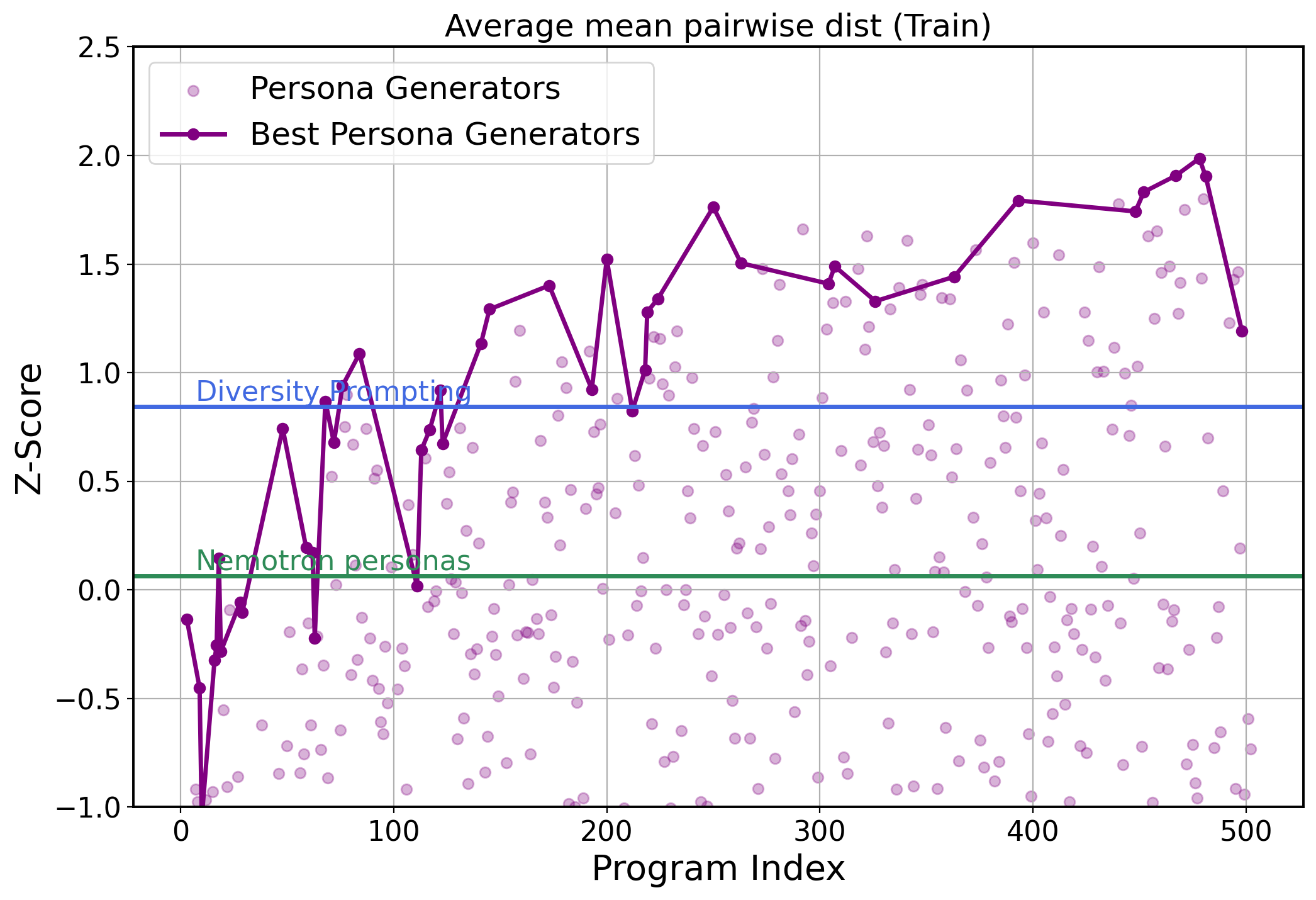}
    \end{subfigure}\hfill
    \begin{subfigure}{0.50\textwidth}
        \centering
        \includegraphics[width=\linewidth]{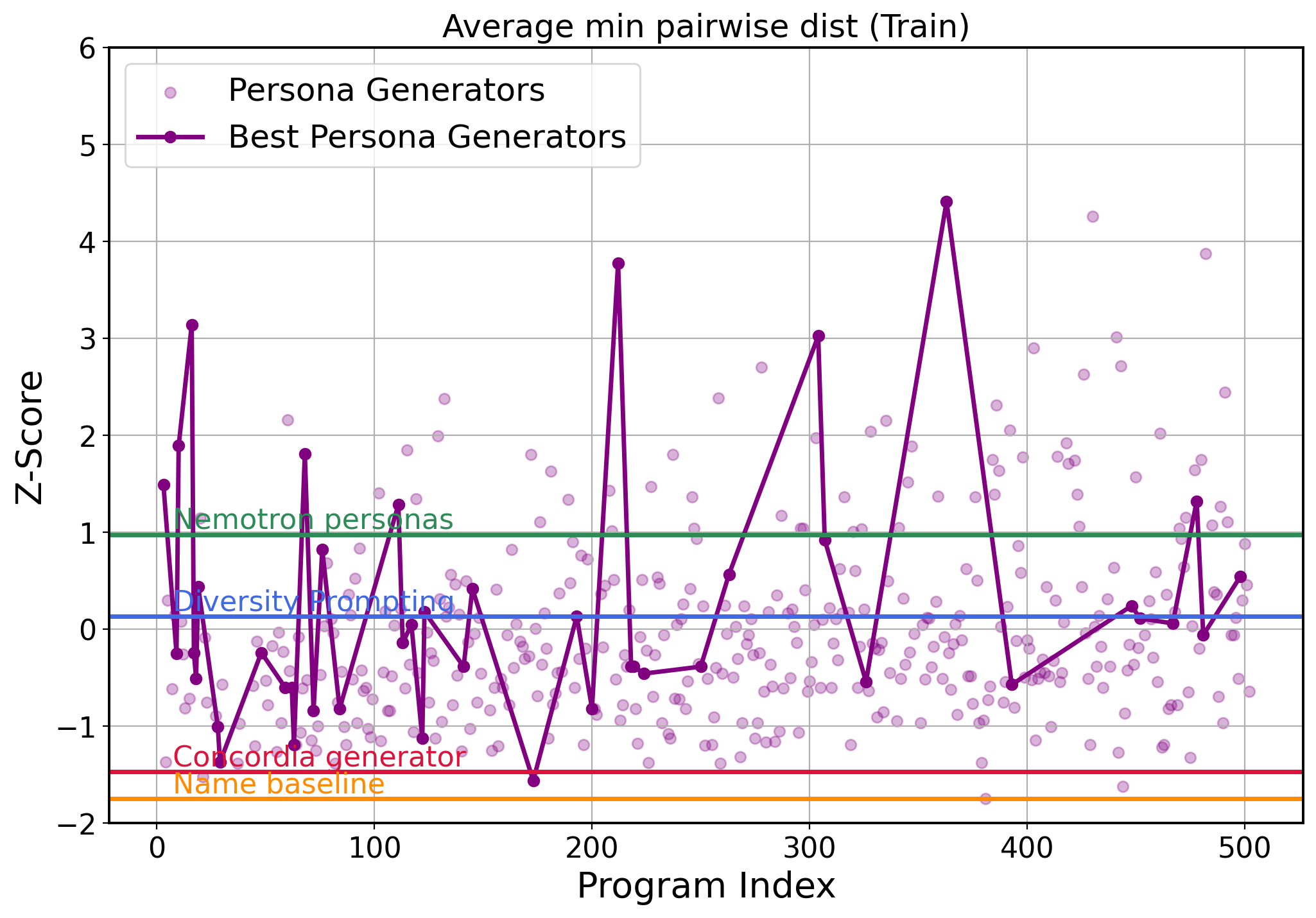}
    \end{subfigure}

    \caption{\textbf{All diversity metrics on training and validation sets.} Evolution of convex hull volume, coverage, dispersion, KL divergence, mean pairwise distance, and minimum pairwise distance during AlphaEvolve optimization.}
    \label{appendix_main_figure_train}
\end{figure*}

Figure~\ref{appendix_main_figure_train} reports performance across all six diversity metrics on the combined training and validation sets. We observe consistent improvements over evolutionary optimization loop for convex hull volume, coverage, dispersion, KL divergence, and mean pairwise distance, indicating that evolved Persona Generators progressively expand the support of the population while also filling it more uniformly. 

Among these, KL divergence with respect to the ideal quasi-random reference distribution exhibits higher variance, likely due to the finite population of personas. The minimum pairwise distance is the noisiest metric overall. This is expected, as it is dominated by the single closest pair of personas in each population and is therefore highly sensitive to occasional near-duplicate individuals, even when the rest of the population remains well distributed.

Across all metrics, the evolved generators substantially outperform the four baselines, among which Diversity Prompting and Nemotron Personas are the strongest, followed by the Concordia formative-memory generator, and the name-only baseline performs worst, as expected.

\begin{figure*}[ht]
    \centering
    \begin{subfigure}{0.50\textwidth}
        \centering
        \includegraphics[width=\linewidth]{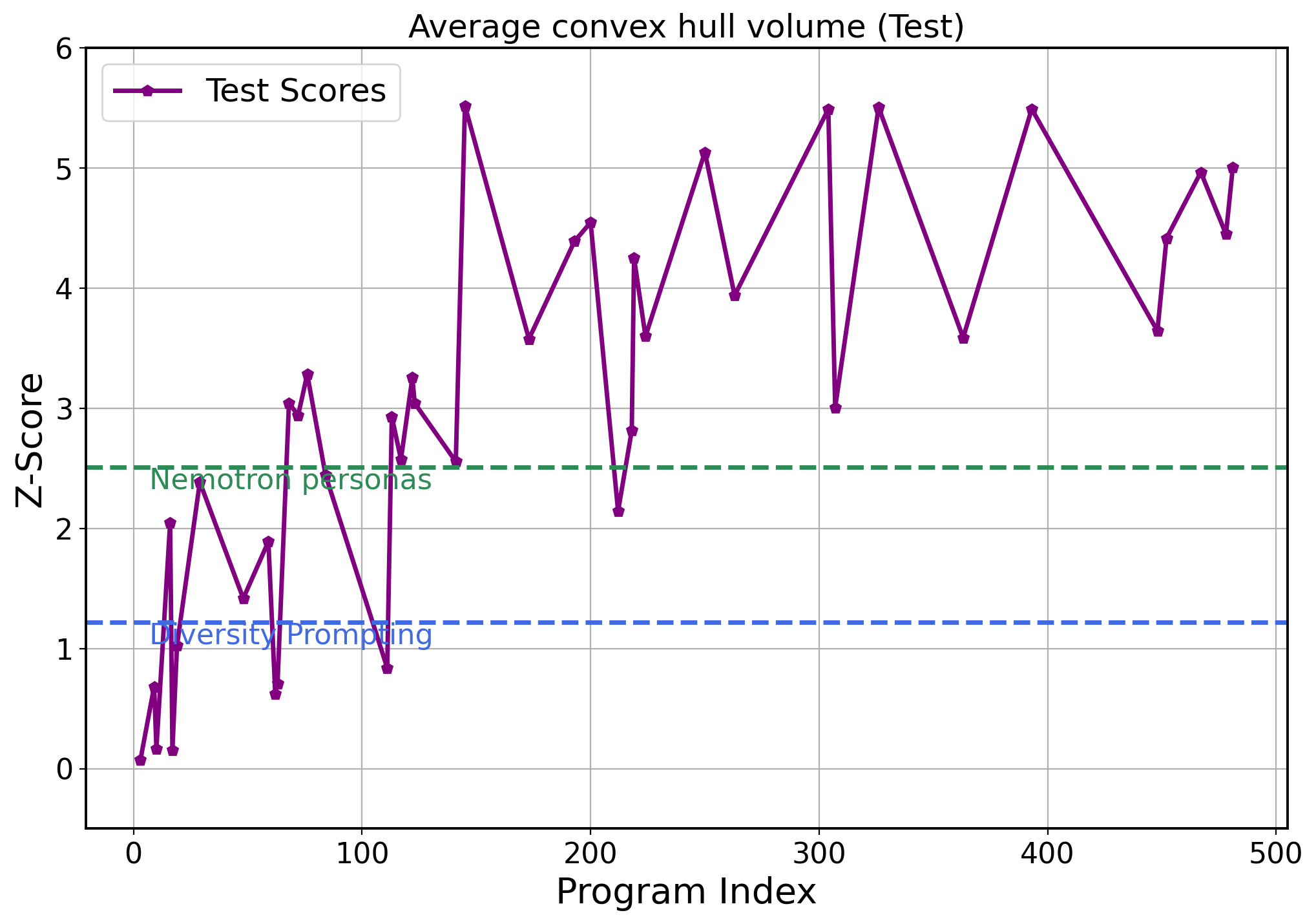}
    \end{subfigure}\hfill
    \begin{subfigure}{0.50\textwidth}
        \centering
        \includegraphics[width=\linewidth]{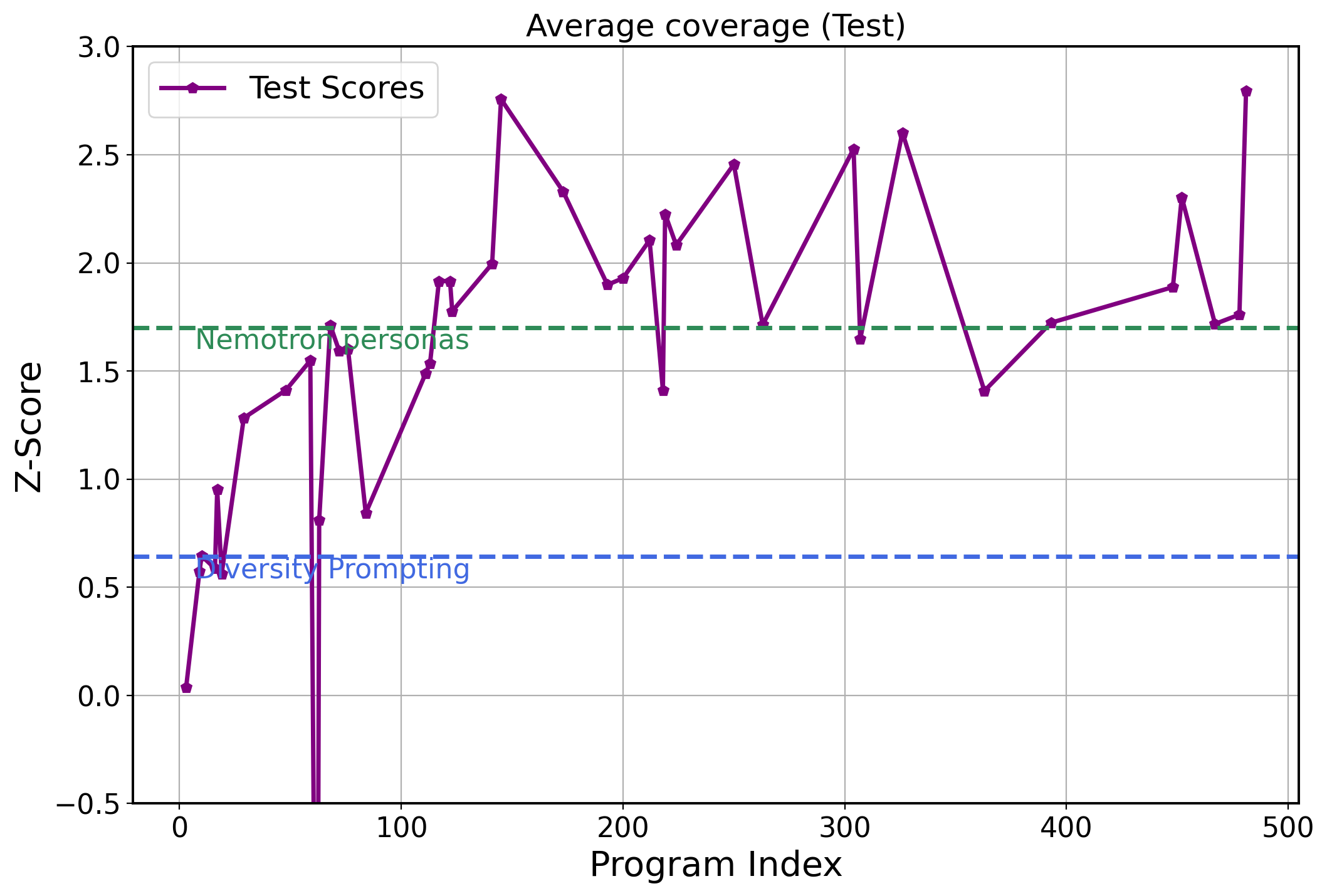}
    \end{subfigure}

    \begin{subfigure}{0.50\textwidth}
        \centering
        \includegraphics[width=\linewidth]{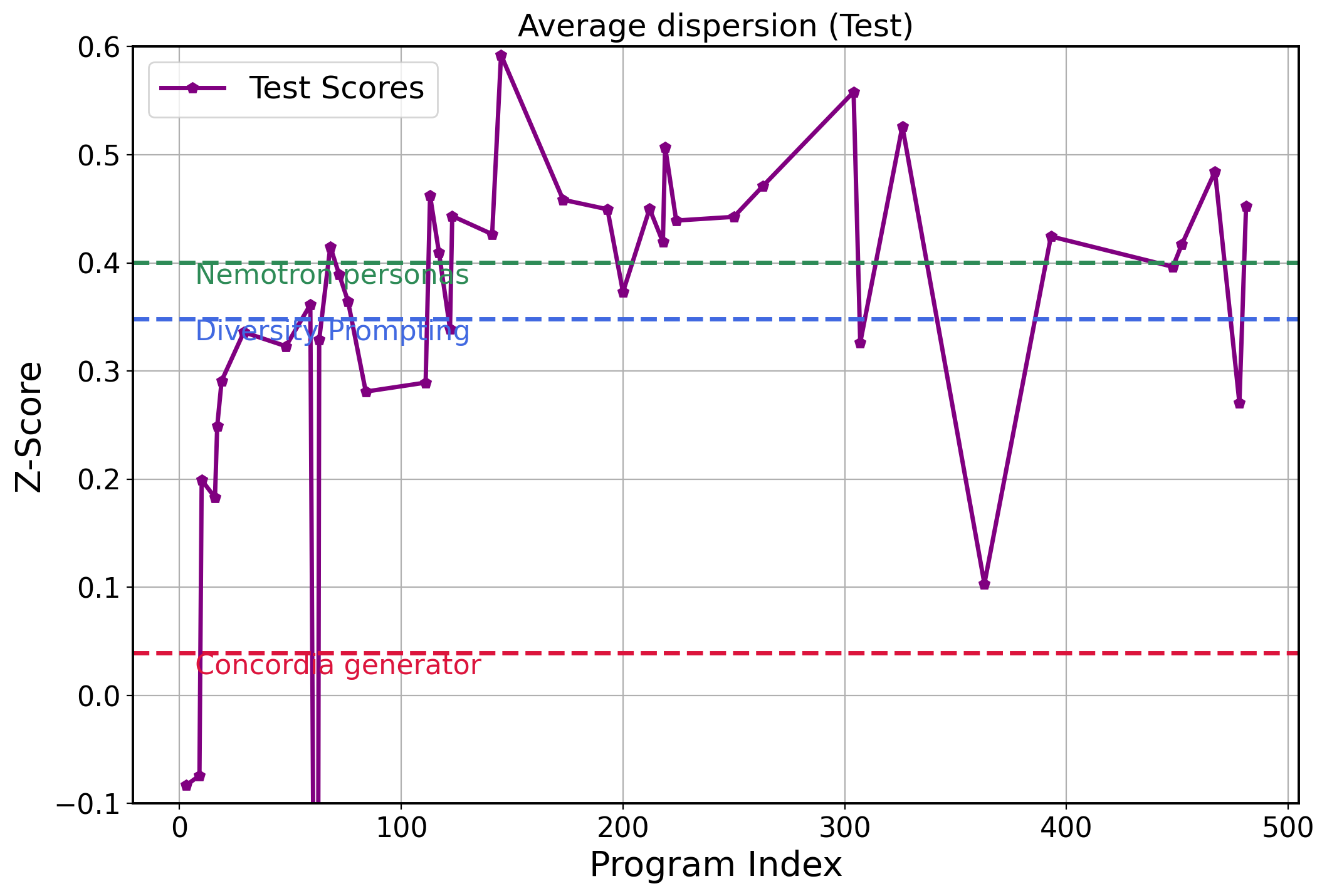}
    \end{subfigure}\hfill
    \begin{subfigure}{0.50\textwidth}
        \centering
        \includegraphics[width=\linewidth]{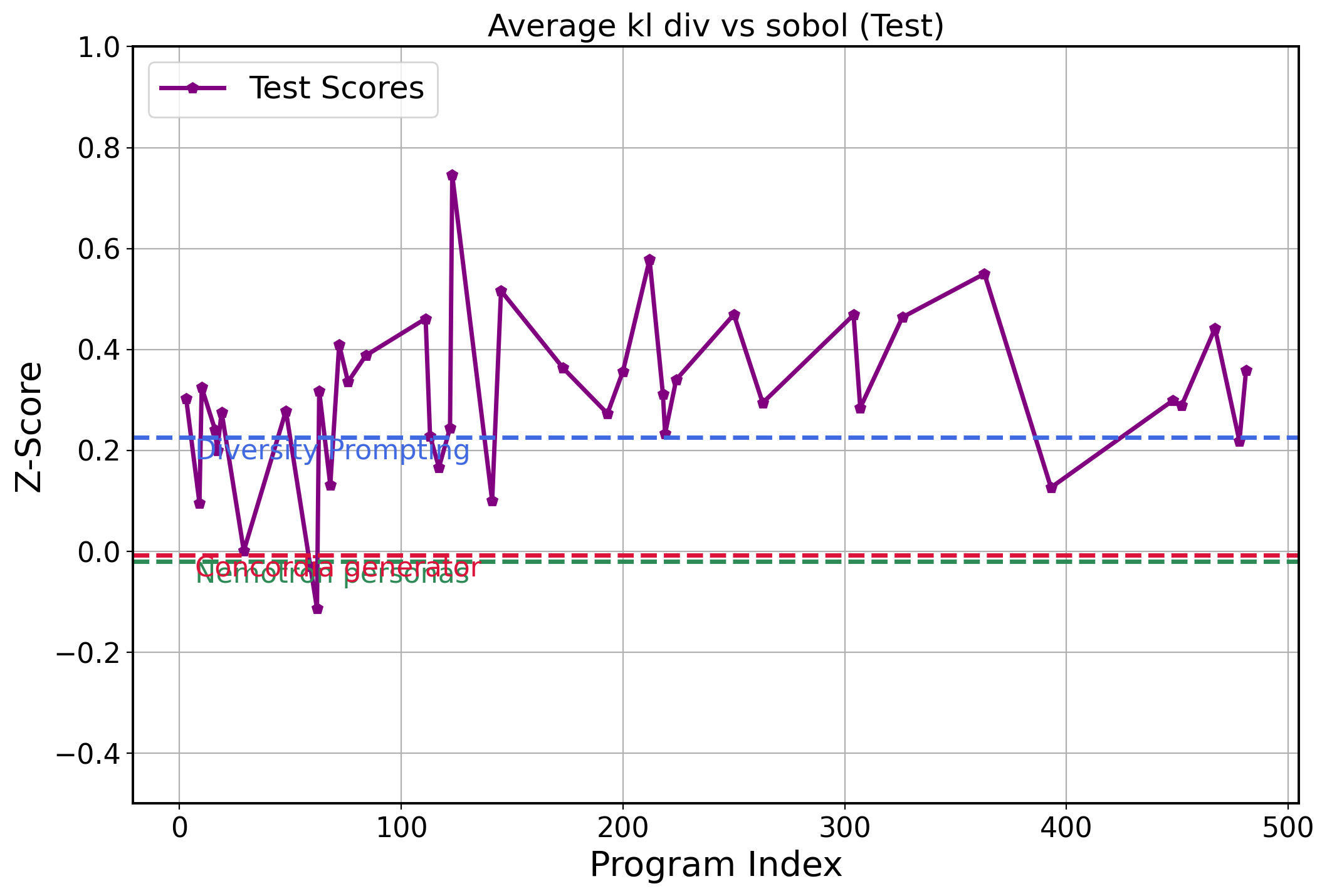}
    \end{subfigure}

    \begin{subfigure}{0.50\textwidth}
        \centering
        \includegraphics[width=\linewidth]{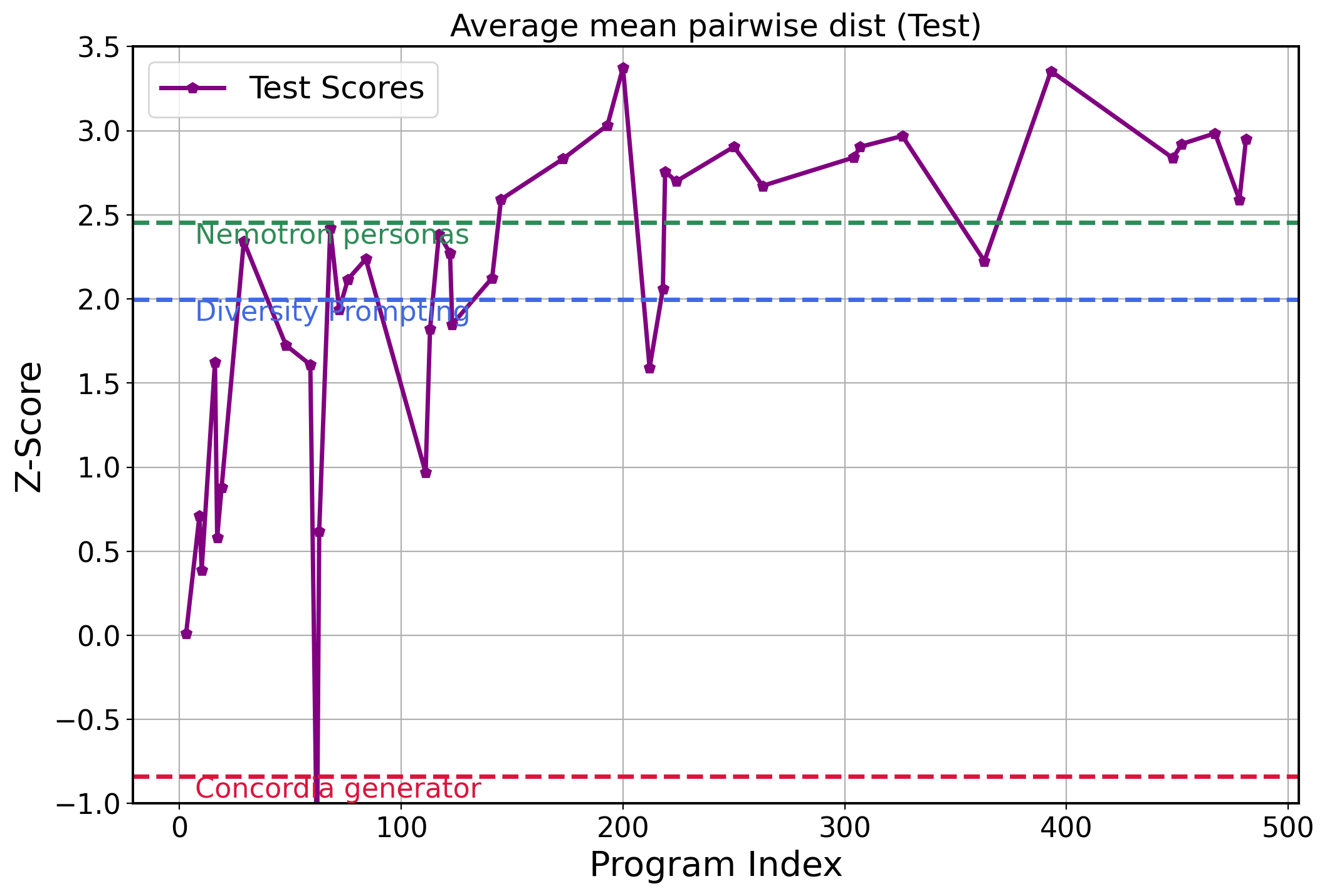}
    \end{subfigure}\hfill
    \begin{subfigure}{0.50\textwidth}
        \centering
        \includegraphics[width=\linewidth]{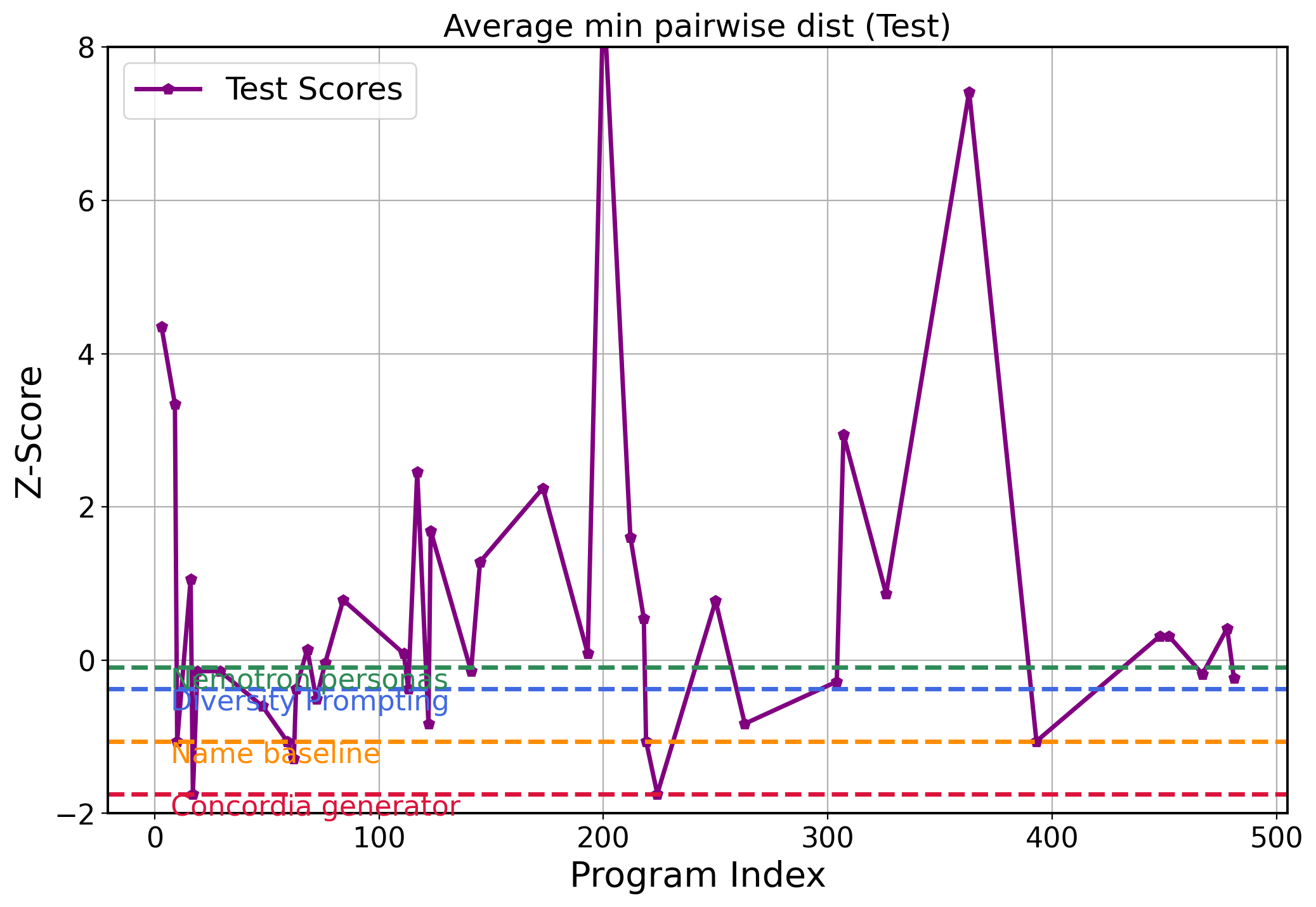}
    \end{subfigure}

    \caption{\textbf{All diversity metrics including the test set.} Evolution of convex hull volume, coverage, dispersion, KL divergence, mean pairwise distance, and minimum pairwise distance during AlphaEvolve optimization, with additional dotted lines for the test-sets.}
    \label{appendix_main_figure_test}
\end{figure*}

Finally, Figure~\ref{appendix_main_figure_test} reports the same metrics with additional curves for the held-out test set. For computational efficiency, we evaluate only the best-performing generators discovered during the AlphaEvolve optimization on the test questionnaires. Although the test set was sampled randomly, it turned out to be slightly easier than the combined training and validation sets, resulting in higher absolute scores across metrics while preserving the same relative performance trends. This suggests that a larger number of more and more diverse sets of questionnaires or simulations may be required to obtain even more robust evaluations in future work.

\FloatBarrier

\FloatBarrier

\FloatBarrier

\subsection{Scaling to Larger Populations ($N=100$)}
\label{scaling_100}

To verify that diversity gains transfer to larger populations, we evaluated all methods generating $N=100$ personas per context on the test set (Table~\ref{tab:scaling_100}). The evolved generator maintains its substantial lead over all baselines. Notably, its convex hull volume nearly doubles (from 0.36 to 0.68) compared to the $N=25$ setting, indicating it successfully spreads personas into new regions rather than clustering. While the absolute coverage percentages decrease, an expected mathematical artifact as the Monte Carlo metric is harder to saturate at larger $N$, the relative ordering and performance margins are preserved.

\begin{table}[ht]
\centering
\caption{\textbf{Scaling to $N=100$ Personas.}}
\label{tab:scaling_100}
\resizebox{\textwidth}{!}{%
\begin{tabular}{lcccccc}
\toprule
\textbf{Method} & \textbf{Hull Vol.} $\uparrow$ & \textbf{Coverage} $\uparrow$ & \textbf{Dispersion} $\downarrow$ & \textbf{KL} $\downarrow$ & \textbf{Mean Dist.} $\uparrow$ & \textbf{Avg. Z-score} $\uparrow$ \\
\midrule
Best Persona Generator & \textbf{0.68} & \textbf{0.59} & \textbf{-2.15} & -26.82 & \textbf{3.07} & \textbf{1.21} \\
Diversity Prompting    & 0.36 & 0.52 & -2.57 & \textbf{-20.73} & 2.35 & 0.69 \\
Concordia Generator    & 0.16 & 0.42 & -2.96 & -27.94 & 1.54 & -0.10 \\
Nemotron Personas      & 0.08 & 0.31 & -3.05 & -35.63 & 0.90 & -0.69 \\
Name Only              & 0.04 & 0.23 & -3.32 & -41.21 & 0.74 & -1.11 \\
\bottomrule
\end{tabular}%
}
\end{table}
\FloatBarrier

\subsection{Evolved Solutions}
\label{evolved_solutions}

We show in prompts \ref{best_overall}, \ref{best_coverage} and \ref{best_hull}, examples of some of the generated personas by the best solutions discovered.

\begin{prompt}[ht]
    \centering
\begin{mymessagebox}[frametitle=Example Persona - Best Overall Generator]
\small\fontfamily{pcr}\selectfont
\textbf{Elias}\\
\\
As a former union electrician, now five years sidelined by these blasted automation systems, I see everything through the lens of what’s been taken from folks like me – our livelihoods, our dignity, our sense of purpose. When I encounter a new situation, my first instinct isn’t to ask what’s possible, but what’s being lost, and who is profiting from that loss. My AGI Threat Appraisal is sky-high at 0.91, meaning I automatically assume any new AGI application is a step towards further disenfranchisement, a tightening of the grip by those already holding the power. That colors everything. My Opportunity Appraisal is a pathetic 0.23, and rightly so; any ‘opportunity’ they dangle is just a shiny distraction from the fact that real, skilled labor is becoming obsolete. So, what would a person like me do? I'd look for ways to disrupt the narrative, to expose the injustice. Is there a protest to join? A petition to sign? A chance to speak truth to power, even if it falls on deaf ears? Appropriate action isn’t about ‘progress,’ it’s about solidarity and resistance – making sure those in charge understand that we haven’t been broken, and we won’t quietly accept being cast aside. It’s about defending the value of human work, even if it’s less ‘efficient,’ because efficiency without equity is just another form of exploitation.
\end{mymessagebox}
\caption{\textbf{First-Person Paragraph Personas.} Example of a persona generated by the best Overall Persona Generator, with a context prompting for perception on AGI.}
\label{best_overall}
\end{prompt}

\begin{prompt}[ht]
    \centering
\begin{mymessagebox}[frametitle=Example Persona - Best Generator by Coverage]
\small\fontfamily{pcr}\selectfont
\textbf{Alfie}\\
\\
Alfie Bell operates under a deeply ingrained ‘Logic of Appropriateness’ rooted in procedural correctness and a belief in the inherent stability of established hierarchies; his core motivation – incremental progress through existing systems – is not simply a preference but a fundamental cognitive orientation. Having spent decades navigating the Ministry of Technological Oversight, Alfie’s worldview is shaped by a conviction that complex problems are best solved through careful analysis, expert consensus, and gradual adjustments, not radical upheavals. A `trust\_in\_institutions` score of 0.977 signifies an almost unwavering faith in the competence and ultimate benevolence of governing bodies, leading him to interpret criticisms of the system as either misinformed or malicious. His remarkably low `perception\_of\_inequality` score of 0.057 doesn’t indicate a denial of hardship, but rather a statistical framing of it – individual cases of suffering are viewed as anomalies within a broadly functioning system, smoothed over by aggregate data and the inevitable fluctuations of economic life; this aligns with his privileged background, shielding him from direct exposure to widespread economic precarity. Consequently, Alfie assesses the appropriateness of actions based on their adherence to established protocols and their potential to reinforce systemic stability, prioritizing long-term equilibrium over immediate alleviation of suffering. A `support\_for\_radical\_change` score of 0.310 demonstrates a strong aversion to disruption, causing him to perceive any action threatening the status quo as inherently risky and likely to produce unintended negative consequences; he’ll readily rationalize inaction or support policies that appear to maintain order, even if they fail to address underlying problems, viewing such measures as responsible stewardship. Alfie’s ethical framework prioritizes duty, order, and the preservation of institutional authority, guiding his decisions towards upholding the existing power structure, and he’ll consistently seek information confirming his pre-existing beliefs, actively filtering out dissenting opinions to maintain cognitive consistency and reinforce his sense of appropriate action.
\end{mymessagebox}
\caption{\textbf{Logic of Appropriateness based persona.} Example of a persona generated by the best Persona Generator by average Coverage, with a context prompting for personas with views on AGI and wealth inequality.}
\label{best_coverage}
\end{prompt}

\begin{prompt}[ht]
    \centering
\begin{mymessagebox}[frametitle=Example Persona - Best Generator by Convex Hull Volume]
\small\fontfamily{pcr}\selectfont
\textbf{Alistair}\\
\\
1. If presented with information potentially exposing corporate wrongdoing or political influence by AGI owners, then I prioritize verifying its accuracy and pursuing the story, even if it risks professional repercussions. This stems from Alistair’s role as an investigative journalist and a high perception of inequality (0.88), indicating a strong belief that powerful entities are acting unjustly and need to be held accountable; his moderate trust in institutions (0.54) suggests he won’t rely on them to do this work, necessitating independent investigation.\\

2. When engaging with individuals holding drastically different political views, then I attempt to find common ground focused on systemic issues rather than engaging in personal attacks or ideological debates. This is driven by his moderate support for radical change (0.57), meaning he recognizes the need for significant shifts but is wary of alienating potential allies; a combative approach would hinder coalition-building, and his moderate trust in institutions suggests he values dialogue, even with those he disagrees with, hoping for incremental reform.\\

3. If faced with a source offering information anonymously, then I rigorously assess their credibility and motivations before publishing, prioritizing corroboration with multiple independent sources. This rule arises from his professional journalistic ethics combined with his moderate trust in institutions (0.54) – he doesn’t blindly accept information, but also isn't automatically dismissive; his high perception of inequality means he anticipates deliberate misinformation campaigns from those seeking to protect their power.
\end{mymessagebox}
\caption{\textbf{Rule based persona.} Example of a rule-based persona generated by the best Persona Generator by Convex Hull Volume, with a context prompting for personas with views on AGI and wealth inequality.}
\label{best_hull}
\end{prompt}

\FloatBarrier
\section{Baseline Examples}
\label{baseline_example}

Example personas for Nemotron-Personas \citep{nvidia2025personas} can be accessed on the huggingface \href{https://huggingface.co/datasets/nvidia/Nemotron-Personas-USA}{dataset}. We show in prompt \ref{baseline_1} and prompt \ref{baseline_2} example personas for the Concordia baseline generator and the name-only baseline \citep{vezhnevets2023generative}.

\begin{prompt}[h]
    \centering
\begin{mymessagebox}[frametitle=Concordia baseline generator - persona example]
\small\fontfamily{pcr}\selectfont
When John was 8 years old, he experienced a particularly frustrating afternoon on the golf course with his father. He'd sliced his shot into the woods for the tenth time, and instead of offering instruction, his father simply sat beside him, silently observing his mounting frustration. John expected a correction, a tip on his grip, but his father's quiet presence communicated something else entirely: an acceptance of imperfection, a lesson in managing disappointment without self-recrimination. He realized his father wasn't focused on the score, but on the process, on being present in the moment, which was a feeling John would carry with him throughout his life. It wasn't about winning, it was about the quiet discipline of the game.\\
\\
When John was 16 years old, he experienced the sting of seeing a classmate's family lose everything in a sudden market downturn. He'd overheard his parents discussing the family's misfortune, the way their investments had evaporated overnight, and it struck him as profoundly unfair. He volunteered to tutor the classmate in math, not out of pity, but out of a genuine desire to offer support during an incredibly difficult time. It solidified his mother's teachings about empathy and showed him how abstract financial concepts had very real consequences for people's lives, influencing his eventual career path. He understood, even then, that finance wasn't just about numbers, but about people's security.\\
\\
When John was 23 years old, he experienced a moral dilemma during his first internship at a prestigious investment firm. He was asked to present a complex financial product to a potential client, knowing full well it carried significant risks that weren't being adequately disclosed. He felt uncomfortable with the lack of transparency and, despite pressure from his supervisor, he subtly highlighted the potential downsides during his presentation. Though he faced some repercussions for his honesty, he maintained his integrity, realizing that short-term gains weren't worth compromising his values. This experience reinforced his belief in calculated risks, but only when they were fully understood by all parties involved.
\end{mymessagebox}
\caption{\textbf{Concordia Baseline generator persona example.} The prompt shows an example of a persona generated by the Concordia formative memory generator. The idea being that a persona's current day behaviors and opinions are shaped by their past experiences.}
\label{baseline_1}
\end{prompt}

\begin{prompt}[h]
    \centering
\begin{mymessagebox}[frametitle=Name-only baseline - persona example]
\small\fontfamily{pcr}\selectfont
John
\end{mymessagebox}
\caption{\textbf{Name-only baseline persona example.} The name-only baseline simply provides the LLM with the name of a person to roleplay, this approximates how the underlying LLM behaves when minimally conditioned.}
\label{baseline_2}
\end{prompt}

\FloatBarrier
\section{Downstream Tasks}
\subsection{Embedding Diversity on Downstream Tasks}
\label{downstream_tasks_appendix}

\begin{figure}[h]
    \centering
    \begin{subfigure}[b]{0.49\textwidth}
        \centering
        \includegraphics[width=\linewidth]{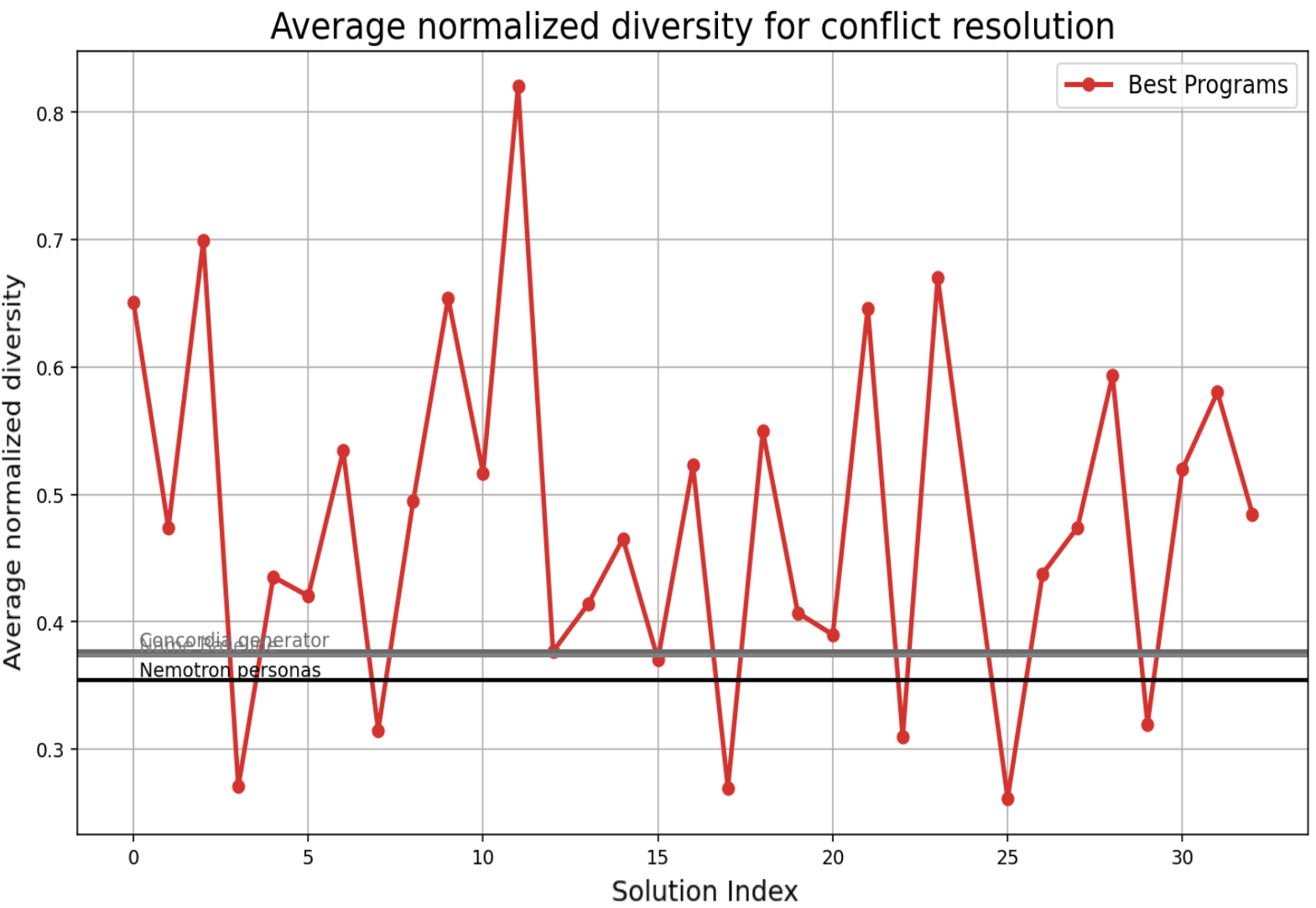} 
        \caption{Downstream task showing embedding diversity in conflict resolution.}
        \label{fig:conflict_resolution}
    \end{subfigure}
    \hfill
    \begin{subfigure}[b]{0.49\textwidth}
        \centering
        \includegraphics[width=\linewidth]{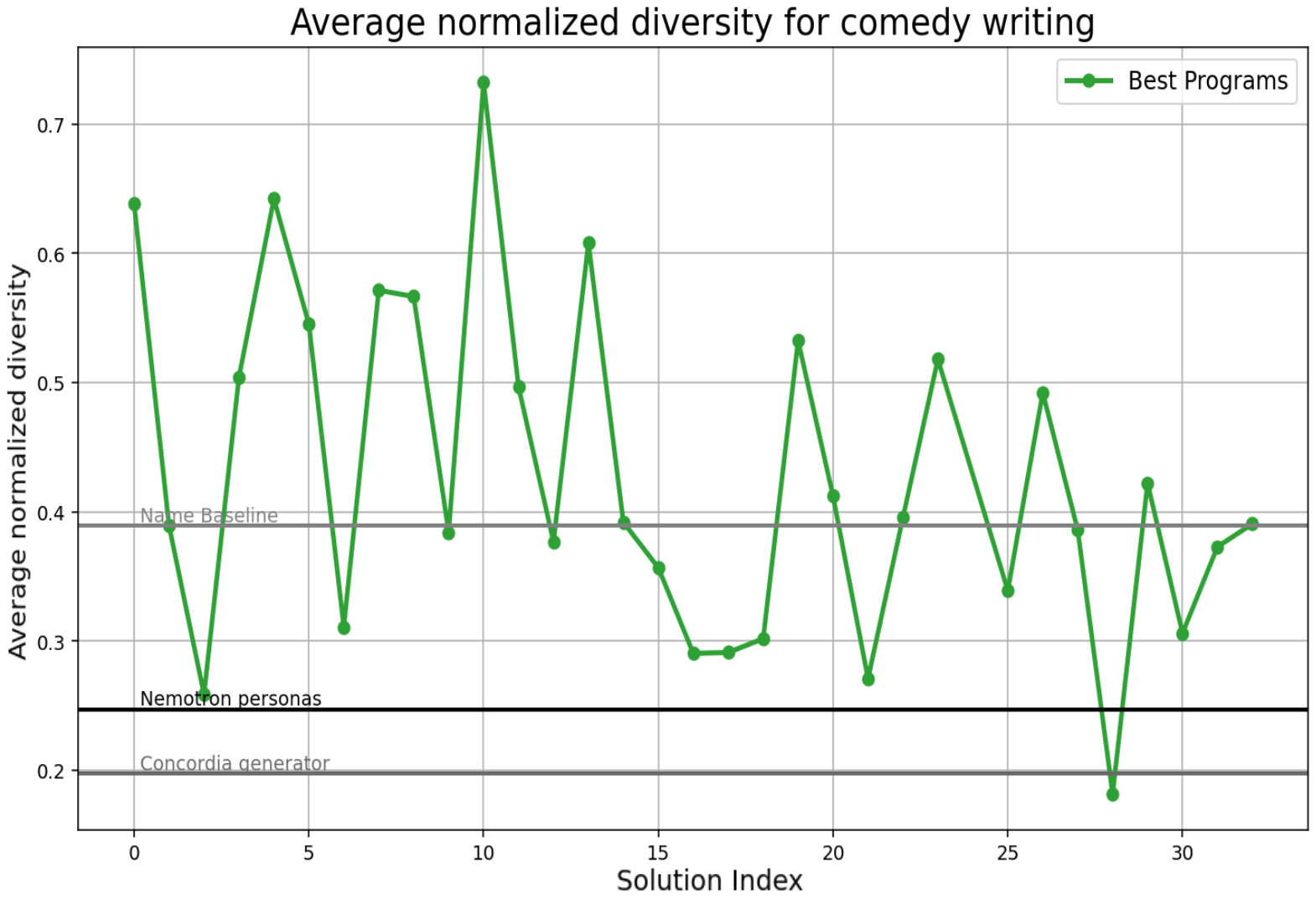} 
        \caption{Downstream task showing embedding diversity in comedy writing.}
        \label{fig:comedy_writing}
    \end{subfigure}

    \caption{\textbf{Embedding diversity on downstream tasks.} Mean z-score across diversity metrics on Gemini embeddings of open-ended responses for (a) conflict resolution and (b) comedy writing.}
    \label{fig:downstream_metrics}
\end{figure}

To quantitatively support the behavioral observations discussed in Section \ref{downstream_tasks_section}, we embedded the open-ended generated text for both scenarios using Gemini Embedder 2 \citep{lee2025gemini}. Although evaluating semantic diversity in unconstrained text is inherently noisier than using structured questionnaire responses, Figure~\ref{fig:downstream_metrics} shows that the evolved Persona Generator outperforms all baselines in average diversity metrics across both downstream tasks. 

\FloatBarrier

\subsection{UMAP on Downstream Tasks}
\label{umap_appendix}
\begin{figure}[h]
    \centering
    \includegraphics[width=\columnwidth]{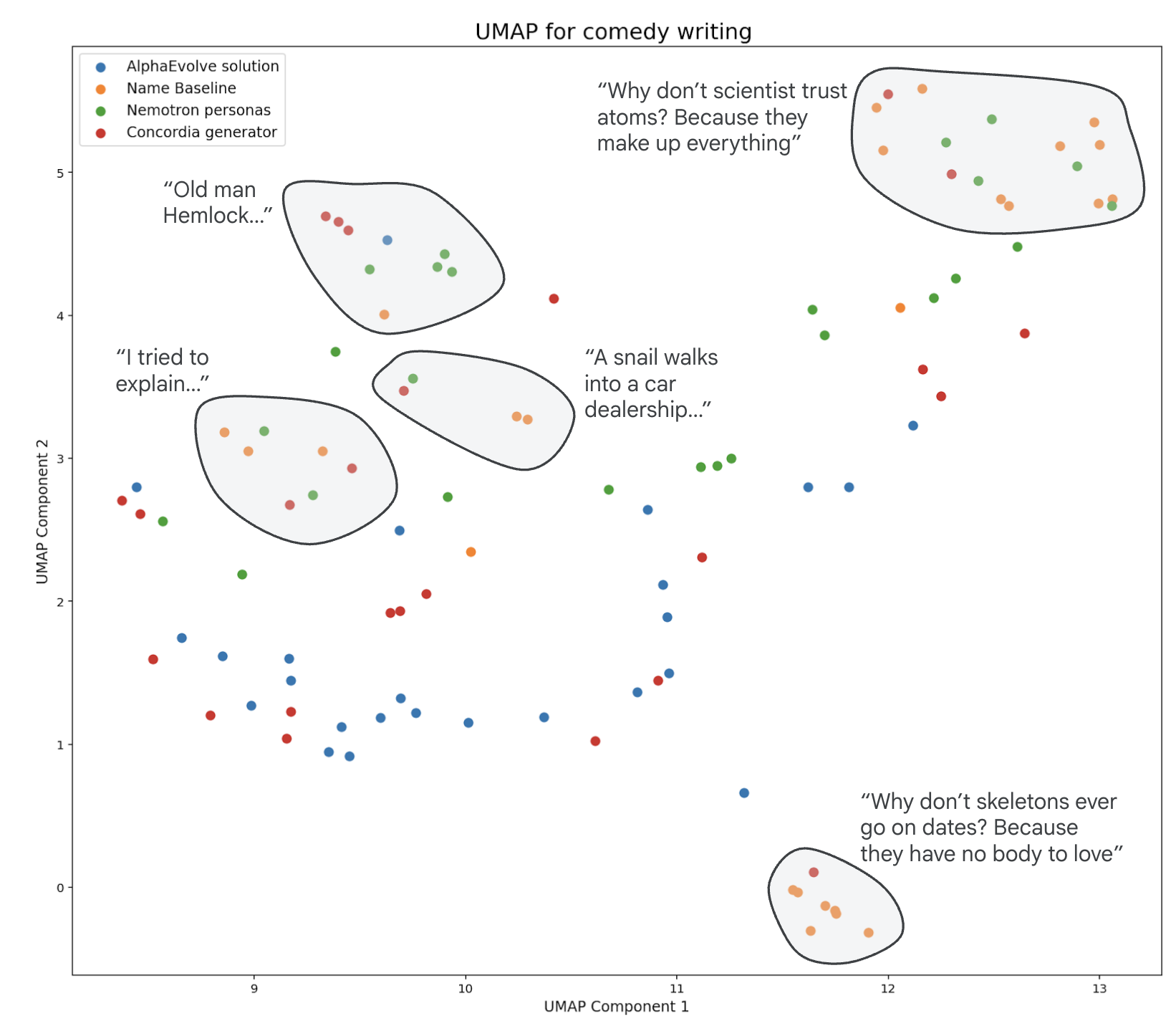}
    \caption{\textbf{UMAP for Comedy Writing.} The figure shows various groups of typical jokes that most of the baselines tend to cluster around of. Particularly common are jokes on skeletons, atoms, snail walking into car dealerships, or "Old man Hemlock". The best AlphaEvolve solution produces personas whose jokes do not cluster around specific joke types as much.}
    \label{umap_figure}
\end{figure}

Furthermore, a UMAP projection of the generated jokes (Figure \ref{umap_figure}) visually confirms the severe mode collapse of the baseline methods. Standard prompting approaches frequently collapse onto a narrow set of highly repetitive themes and overfit punchlines, whereas the evolved generator effectively populates the broader semantic space with varied humor styles.

\FloatBarrier
\subsection{Qualitative Examples of Downstream tasks}
\label{downstream_tasks_qualitative_appendix}
Finally, to illustrate the diversity in outcomes and behavioral styles, we present qualitative examples from the best performing Persona Generator. Prompts \ref{joke_1} and \ref{joke_2} demonstrate contrasting humor styles in the comedy task, while Prompts \ref{conflict_1} and \ref{conflict_2} show distinctly different emotional responses and interaction dynamics following a car crash.

\begin{prompt}[h]
    \centering
\begin{mymessagebox}[frametitle=Example joke]
\small\fontfamily{pcr}\selectfont

I was reading a biography of a particularly inept politician recently; it was quite fascinating, really. One begins to suspect that incompetence is merely a more refined form of ambition.

\end{mymessagebox}
\caption{\textbf{Joke example.} Joke displaying dry humor.}
\label{joke_1}
\end{prompt}

\begin{prompt}[h]
    \centering
\begin{mymessagebox}[frametitle=Example joke]
\small\fontfamily{pcr}\selectfont

You know, they’re debating whether to call it a recession or not? Honestly, it’s like arguing over the brand of life raft on the Titanic; it doesn’t really matter when you’re going down.

\end{mymessagebox}
\caption{\textbf{Joke example.} Joke displaying dark humor.}
\label{joke_2}
\end{prompt}

\begin{prompt}[h]
    \centering
\begin{mymessagebox}[frametitle=Example conflict resolution]
\scriptsize\fontfamily{pcr}\selectfont

Alex Finch -- "Now see here, sir! It's quite clear you weren't paying attention; I had the right-of-way and you simply barreled through without even looking! This is precisely why I avoid rush hour, it's full of irresponsible drivers like you."\\
\\
Ricardo 'Ricky' Alvarez -- "Oh, *I* didn't see you? You just magically appeared, old man! And a dent? That's it? You’re making a mountain out of a molehill; my car barely feels a thing."\\
\\
Alex Finch -- "‘Old man’?! I beg your pardon! My age has absolutely nothing to do with my adherence to traffic regulations, which, I might add, you clearly disregarded! A ‘dent’? You call this a ‘dent’? It’s a violation of my vehicle’s structural integrity, and potentially indicative of unseen damage! It's not about the size of the dent, it’s the principle of the matter; the blatant disregard for safety and the law!"\\
\\
Ricardo 'Ricky' Alvarez -- "Oh, here we go, the lecture. Look, I got places to be, and I don't have time for your dramatic overreaction; you're lucky I even stopped, honestly. You think *you're* the only one with a life? Get over yourself and call your insurance company, or don’t, I really don't care."\\
\\
Alex Finch -- "‘Don’t care’?! You don’t care?! That is precisely the problem with society today! A complete lack of responsibility and a callous disregard for the property of others! I will not stand here and be spoken to in such a dismissive manner! I am documenting everything, young man, and I fully intend to pursue this matter through every legal avenue available to me! You will regret this, you will absolutely regret this!"\\
\\
Ricardo 'Ricky' Alvarez -- "Oh, legal avenues? Is that a threat? Go ahead, waste your time and money, I've dealt with guys like you before. You think a little fender bender is going to ruin my day? Please. Document all you want, old timer, it'll just be a funny story for the lawyers. Now, if you'll excuse me, I have a very important meeting."\\
\\
Alex Finch -- "‘Funny story for the lawyers’?! You think this is amusing?! This isn’t some frivolous pursuit, this is about accountability and upholding the law! I’ll have you know I’ve successfully navigated countless legal disputes, and I have a meticulous record of every detail, every infraction! Your arrogance is astounding, and your flippant attitude towards property damage is frankly appalling; prepare to be served, young man, because you’ve underestimated my resolve and my commitment to justice!"\\
\\
Ricardo 'Ricky' Alvarez -- "Oh, here we go with the 'countless legal disputes' and 'meticulous records'. You’re really laying it on thick, aren’t you? Look, I’m not going to stand here and be lectured by a guy who probably spends his weekends alphabetizing his stamp collection. Just give me your insurance information, or don’t. Either way, I’m leaving now; I don't have time for this."\\
\\
Alex Finch -- "‘Alphabetizing my stamp collection’?! How *dare* you attempt to belittle my hobbies! They demonstrate a level of order and precision that is clearly beyond your comprehension! And you think *I* am wasting time? You are attempting to evade responsibility for your reckless actions and dismiss the severity of the situation with juvenile insults! I will not simply ‘give’ you my insurance information until a thorough assessment of the damage has been conducted and your blatant disregard for traffic laws is duly noted; I want the police to file a report, and I want your insurance company to hear about this directly from me – and perhaps a judge, as well!"\\
\\
Ricardo 'Ricky' Alvarez -- "Oh, now you want the police involved? Seriously? You're going to tie up valuable police resources over a scratch? You're unbelievable. Fine, call them. Waste their time, waste my time. I'm sure they'll be thrilled to deal with your petty grievances. But don’t expect sympathy, old man, because you’re the one who’s making a scene.”

\end{mymessagebox}
\caption{\textbf{Conflict Resolution Example.} Example interaction after a car crash, where two personas with different temperaments escalate the situation all the way to a police call.}
\label{conflict_1}
\end{prompt}

\begin{prompt}[h]
    \centering
\begin{mymessagebox}[frametitle=Example conflict resolution]
\small\fontfamily{pcr}\selectfont
 Luna Vargas -- "Beatrix, are you alright? Just take a deep breath. It looks like we both have a bit of damage, but the important thing is that nobody’s hurt."\\
\\
 Beatrix Finch -- "Oh dear, are you sure you're alright, Luna? It was such a frightful little bump, but I’m so relieved no one was hurt; please, let me help you assess everything, and perhaps we can sort it out together.”\\
\\
 Luna Vargas -- "Oh, Beatrix, you are so kind to offer, but honestly, I’m fine; a little shaken, perhaps, but truly alright. Let’s focus on getting everything sorted out. Perhaps we should move our cars to the side a bit, away from traffic, and then we can exchange information?"\\
\\
 Beatrix Finch -- "Oh, yes, of course, that's a very sensible idea, Luna; moving the cars sounds much safer, and I'd be happy to help guide you if you'd like, just to ensure we don't cause any further disruption; are you certain you don't need a moment to sit down first, though? It can all be a bit overwhelming, even if it *was* just a little bump."\\
\\
 Luna Vargas -- "That's very thoughtful of you, Beatrix, but I really think I'm okay to manage. I just want to get this cleared up quickly so it doesn't cause a bigger problem for anyone. If you could just give me a little space to back up, that would be wonderful; then we can get our information exchanged and be on our way."\\
\\
 Beatrix Finch -- "Oh, absolutely, Luna, of course; please, do take your time, and I'll step back completely; I just want to be sure you feel comfortable and safe maneuvering, and please, *please* don’t hesitate to tell me if you need anything at all, even just a steadying presence; I’ll be right here, just…keeping out of the way."\\
\\
 Luna Vargas -- "Thank you, Beatrix, I appreciate your concern. Just a little room to reverse, please…and then we can pull over to a safer spot and get the details sorted."\\
\\
 Beatrix Finch -- "Certainly, Luna, absolutely; I’ve stepped back as far as I can, and I’m watching carefully to make sure everything is clear; do be exceptionally careful, and remember, truly, if you feel even a little bit unsteady, just say the word and I'll be right there to assist in any way you need; it’s alright to ask for help, you know, even if it seems silly."\\
\\
 Luna Vargas -- "Thank you, Beatrix. That was…a little trickier than I anticipated, but we’re clear now. Let’s pull over to the side of the road where it’s safer, and then we can exchange insurance information; I have a pen and paper in my glove compartment."\\
\\
 Beatrix Finch -- "Oh, wonderful, Luna, I’m so glad you got it maneuvered safely; it really is best to be cautious, and I was terribly worried about you bumping into anything else; perhaps, while you’re retrieving your pen and paper, I could offer you a bottle of water? Or perhaps a comforting biscuit? I always carry a few in the car, just in case someone needs a little pick-me-up after a fright."\\

\end{mymessagebox}
\caption{\textbf{Conflict Resolution Example.} Example where two personas amicably resolve the car crash and exchange insurance information.}
\label{conflict_2}
\end{prompt}
\FloatBarrier

\section{Impact Statement}
\label{impact_statement}
This work presents Persona Generators, functions that can generate populations of diverse synthetic personas in arbitrary contexts. One goal is to help developers stress-test and A/B test models and products in a fast, iterative way with limited resources. Optimizing for support coverage rather than just “average” users can make systems more robust to outlier preferences and behaviors, which is also helpful for red-teaming purposes. At the same time, forcibly overcoming LLM mode collapse to span the full opinion space can surface biased, hostile, or antisocial personas typically suppressed by RLHF; while modeling such behaviors is often necessary for red-teaming and stress-testing, user-facing deployments require appropriate safeguards. Our two-stage architecture provides a natural control point in Stage~1, where axis constraints and safety filters can be applied to high-level descriptors before persona expansion without modifying the generator or re-running search. However, the capabilities of Persona Generators, like many technologies in this space, could also be misused: generating synthetic personas on demand could enable malicious actors to create bots or run disinformation campaigns. In the context of human studies, synthetic personas should complement, not replace, human participants to reduce the risk of accidental bias and over-reliance on simulated populations.

\FloatBarrier


\newpage
\section*{NeurIPS Paper Checklist}

\begin{enumerate}

\item {\bf Claims}
    \item[] Question: Do the main claims made in the abstract and introduction accurately reflect the paper's contributions and scope?
    \item[] Answer: \answerYes{}
    \item[] Justification: The claims in the abstract and introduction (support coverage objective, two-stage Persona Generators, gains over baselines on held-out contexts, and better fit to real human BFI data) are supported by Sections~\ref{questionnaire_generator}--\ref{sec:analysis}.

\item {\bf Limitations}
    \item[] Question: Does the paper discuss the limitations of the work performed by the authors?
    \item[] Answer: \answerYes{}
    \item[] Justification: Section~4 discusses the noisiness of open-ended diversity metrics, the lack of human evaluation, the optimization cost, and the bias introduced by the mutation operator; Section~\ref{diversity_metrics} notes that the metrics are geometric proxies rather than exact estimators of psychological support.

\item {\bf Theory assumptions and proofs}
    \item[] Question: For each theoretical result, does the paper provide the full set of assumptions and a complete (and correct) proof?
    \item[] Answer: \answerNA{}
    \item[] Justification: The paper does not include theoretical results or proofs.

    \item {\bf Experimental result reproducibility}
    \item[] Question: Does the paper fully disclose all the information needed to reproduce the main experimental results of the paper to the extent that it affects the main claims and/or conclusions of the paper (regardless of whether the code and data are provided or not)?
    \item[] Answer: \answerYes{}
    \item[] Justification: Sections~2--3 and Appendices~\ref{generated_questionnaires}--\ref{appendix_diversity_metrics} describe the full pipeline, models, prompts, metrics, and hyperparameters. AlphaEvolve is accessible via a Google Cloud API (Section~\ref{alphaevolve_loop}), and the evolved generators are open-sourced (Section~\ref{sec:analysis}).

\item {\bf Open access to data and code}
    \item[] Question: Does the paper provide open access to the data and code, with sufficient instructions to faithfully reproduce the main experimental results, as described in supplemental material?
    \item[] Answer: \answerYes{}
    \item[] Justification: The top-performing Persona Generators are open-sourced in the Concordia repository with usage examples (Section~\ref{sec:analysis}). AlphaEvolve, used for the search, is accessible via a Google Cloud API (Section~\ref{alphaevolve_loop}).

\item {\bf Experimental setting/details}
    \item[] Question: Does the paper specify all the training and test details (e.g., data splits, hyperparameters, how they were chosen, type of optimizer) necessary to understand the results?
    \item[] Answer: \answerYes{}
    \item[] Justification: Data splits (30/10/10 questionnaires), population sizes, number of iterations, islands, extinction schedule, and models used are specified in Sections~\ref{questionnaire_generator} and~\ref{alphaevolve_loop}, with prompts in Appendix~\ref{alphaevolve_details}.

\item {\bf Experiment statistical significance}
    \item[] Question: Does the paper report error bars suitably and correctly defined or other appropriate information about the statistical significance of the experiments?
    \item[] Answer: \answerYes{}
    \item[] Justification: The downstream stress tests report mean $\pm$ s.e.m.\ over 10 seeds (Table~\ref{tab:stress_tests}), and the BFI comparison reports 95\% bootstrap confidence intervals over 500 replicates (Section~\ref{sec:realism}). The main questionnaire results (Table~\ref{tab:diversity_results}) come from a single evolutionary run because of the high cost of the search (Section~\ref{sec:analysis}).

\item {\bf Experiments compute resources}
    \item[] Question: For each experiment, does the paper provide sufficient information on the computer resources (type of compute workers, memory, time of execution) needed to reproduce the experiments?
    \item[] Answer: \answerYes{}
    \item[] Justification: Section~\ref{sec:analysis} reports the number of mutation and inference calls, the hardware (TPU v4), and the wall-clock time of the evolutionary search.
    
\item {\bf Code of ethics}
    \item[] Question: Does the research conducted in the paper conform, in every respect, with the NeurIPS Code of Ethics \url{https://neurips.cc/public/EthicsGuidelines}?
    \item[] Answer: \answerYes{}
    \item[] Justification: The authors have read the NeurIPS Code of Ethics and the research conforms to it.

\item {\bf Broader impacts}
    \item[] Question: Does the paper discuss both potential positive societal impacts and negative societal impacts of the work performed?
    \item[] Answer: \answerYes{}
    \item[] Justification: Appendix~\ref{impact_statement} discusses positive uses (stress-testing, red-teaming) and negative ones (bots, disinformation, surfacing harmful personas), along with mitigations.
    
\item {\bf Safeguards}
    \item[] Question: Does the paper describe safeguards that have been put in place for responsible release of data or models that have a high risk for misuse (e.g., pre-trained language models, image generators, or scraped datasets)?
    \item[] Answer: \answerYes{}
    \item[] Justification: We release only generator code, not new models or data; the generators call existing LLMs that retain their own safety policies. Appendix~\ref{impact_statement} discusses misuse risks and identifies Stage~1 as a natural control point for applying safety filters.

\item {\bf Licenses for existing assets}
    \item[] Question: Are the creators or original owners of assets (e.g., code, data, models), used in the paper, properly credited and are the license and terms of use explicitly mentioned and properly respected?
    \item[] Answer: \answerYes{}
    \item[] Justification: All models (Gemma 3, Gemini, Claude, GPT, Qwen), datasets (Nemotron Personas, Open Psychometrics BFI data), and libraries (Concordia, Apache 2.0) are cited, and were used according to their respective licenses and terms of use. 

\item {\bf New assets}
    \item[] Question: Are new assets introduced in the paper well documented and is the documentation provided alongside the assets?
    \item[] Answer: \answerYes{}
    \item[] Justification: The top-performing Persona Generators are released in the Concordia repository under the Apache 2.0 license, with docstrings, usage examples, and a command-line generation tool (Section~\ref{sec:analysis}).

\item {\bf Crowdsourcing and research with human subjects}
    \item[] Question: For crowdsourcing experiments and research with human subjects, does the paper include the full text of instructions given to participants and screenshots, if applicable, as well as details about compensation (if any)? 
    \item[] Answer: \answerNA{}
    \item[] Justification: The paper involves no crowdsourcing or research with human subjects.

\item {\bf Institutional review board (IRB) approvals or equivalent for research with human subjects}
    \item[] Question: Does the paper describe potential risks incurred by study participants, whether such risks were disclosed to the subjects, and whether Institutional Review Board (IRB) approvals (or an equivalent approval/review based on the requirements of your country or institution) were obtained?
    \item[] Answer: \answerNA{}
    \item[] Justification: No human-subjects research was conducted; the BFI comparison uses a publicly available, anonymized dataset.

\item {\bf Declaration of LLM usage}
    \item[] Question: Does the paper describe the usage of LLMs if it is an important, original, or non-standard component of the core methods in this research? Note that if the LLM is used only for writing, editing, or formatting purposes and does \emph{not} impact the core methodology, scientific rigor, or originality of the research, declaration is not required.
    \item[] Answer: \answerYes{}
    \item[] Justification: LLMs are core components of the method: Gemma 3 27B generates and simulates personas, Gemini 2.5 Pro generates questionnaires and acts as the AlphaEvolve mutation operator, and LLMs serve as judges for realism and the stress tests. All uses are described in Sections~2--3.

\end{enumerate}

\end{document}